%% file: main.tex
\documentclass[11pt]{article}
\pdfoutput=1

\usepackage{makecell}

\usepackage[]{acl}

\usepackage{times}
\usepackage{latexsym}

\usepackage[T1]{fontenc}

\usepackage[utf8]{inputenc}
\usepackage[namelimits]{amsmath} 

\usepackage{microtype}

\usepackage{subcaption}
\usepackage{inconsolata}
\usepackage{algorithmic}
\usepackage{algorithm}
\usepackage{colortbl}
\definecolor{mycyan}{cmyk}{.2,0,0,0}
\usepackage{url}
\usepackage{graphicx}
\usepackage{multirow}
\usepackage{booktabs} 
\usepackage{longtable}
\usepackage{tcolorbox}
\usepackage{hyperref}
\usepackage{lipsum}
\usepackage{footnote}

\newcommand{\eat}[1]{}
\def\hjw{\textcolor{black}}

\def\wqyr{\textcolor{black}}
\newcommand{\methodname}{DOME}

\title{Generating Long-form Story Using Dynamic Hierarchical \\Outlining with Memory-Enhancement}

\author{
    Qianyue Wang \textsuperscript{\rm 1 \rm 2}\footnotemark[1]~
    Jinwu Hu \textsuperscript{\rm 1 \rm 2}\footnotemark[1]~
    Zhengping Li \textsuperscript{\rm 1}~
    Yufeng Wang \textsuperscript{\rm 1 \rm 3}\\
    \textbf{Daiyuan Li} \textsuperscript{\rm 1}~
    \textbf{Yu Hu} \textsuperscript{\rm 4}~ 
    \textbf{Mingkui Tan} \textsuperscript{\rm 1} \footnotemark[2] ~\\ 
    \textsuperscript{\scriptsize{\rm 1}}\small{South China University of Technology,}~
    \textsuperscript{\rm 2}\small{Pazhou Laboratory,}~
    \textsuperscript{\rm 3}\small{Peng Cheng Laboratory,}~ \\
    \textsuperscript{\rm 4}\small{Hong Kong Polytechnic University}\\
}

\begin{document}
\maketitle
\footnotetext[1]{Equal contribution: 202420145083@scut.mail.edu.cn, fhujinwu@gmail.com} 
\footnotetext[2]{Corresponding author: mingkuitan@scut.edu.cn}

\begin{abstract}
\hjw{Long-form story generation task aims to produce coherent and sufficiently lengthy text, essential for applications such as novel writing and interactive storytelling. \wqyr{However, existing methods, including LLMs, rely on rigid outlines or lack macro-level planning, making it difficult to achieve both contextual consistency and coherent plot development in long-form story generation.} To address \wqyr{this} issues, we propose \textbf{D}ynamic Hierarchical \textbf{O}utlining with \textbf{M}emory-\textbf{E}nhancement long-form story generation method, named \textbf{DOME}, to generate the long-form story with coherent content and plot. Specifically, the Dynamic Hierarchical Outline (DHO) mechanism incorporates the novel writing theory into outline planning and fuses the plan and writing stages together, improving the coherence of the plot by ensuring the plot completeness and \wqyr{adapting to the uncertainty during story generation. }
A Memory-Enhancement Module (MEM) based on temporal knowledge graphs is introduced to store and access \wqyr{the} generated content, reducing contextual conflicts and improving story coherence. Finally, we propose a Temporal Conflict Analyzer leveraging temporal knowledge graphs to \wqyr{automatically} evaluate the contextual consistency of long-form story. 
Experiments demonstrate that DOME significantly improves the fluency, coherence, and overall quality of generated long stories compared to state-of-the-art methods.}
\end{abstract}

\input{section/1_Introduction}

\input{section/2_RelatedWork}

\input{section/Problem_definition}

\input{section/3_Method}

\input{section/4_Experiment}

\input{section/5_Analysis}

\newpage

\input{section/6_Limitation}

\section*{Ethics Statement}
\hjw{This work fully complies with the \href{https://www.aclweb.org/portal/content/acl-code-ethics}{ACL Ethics Policy}. We declare that there are no ethical issues in this paper, to the best of our knowledge.}

\bibliography{mainbib}
\newpage
\input{section/Appendix}

\end{document}

%% file: section/1_Introduction.tex
\section{Introduction}

\begin{figure}[t]  
    \centering  
    \includegraphics[scale=0.36]{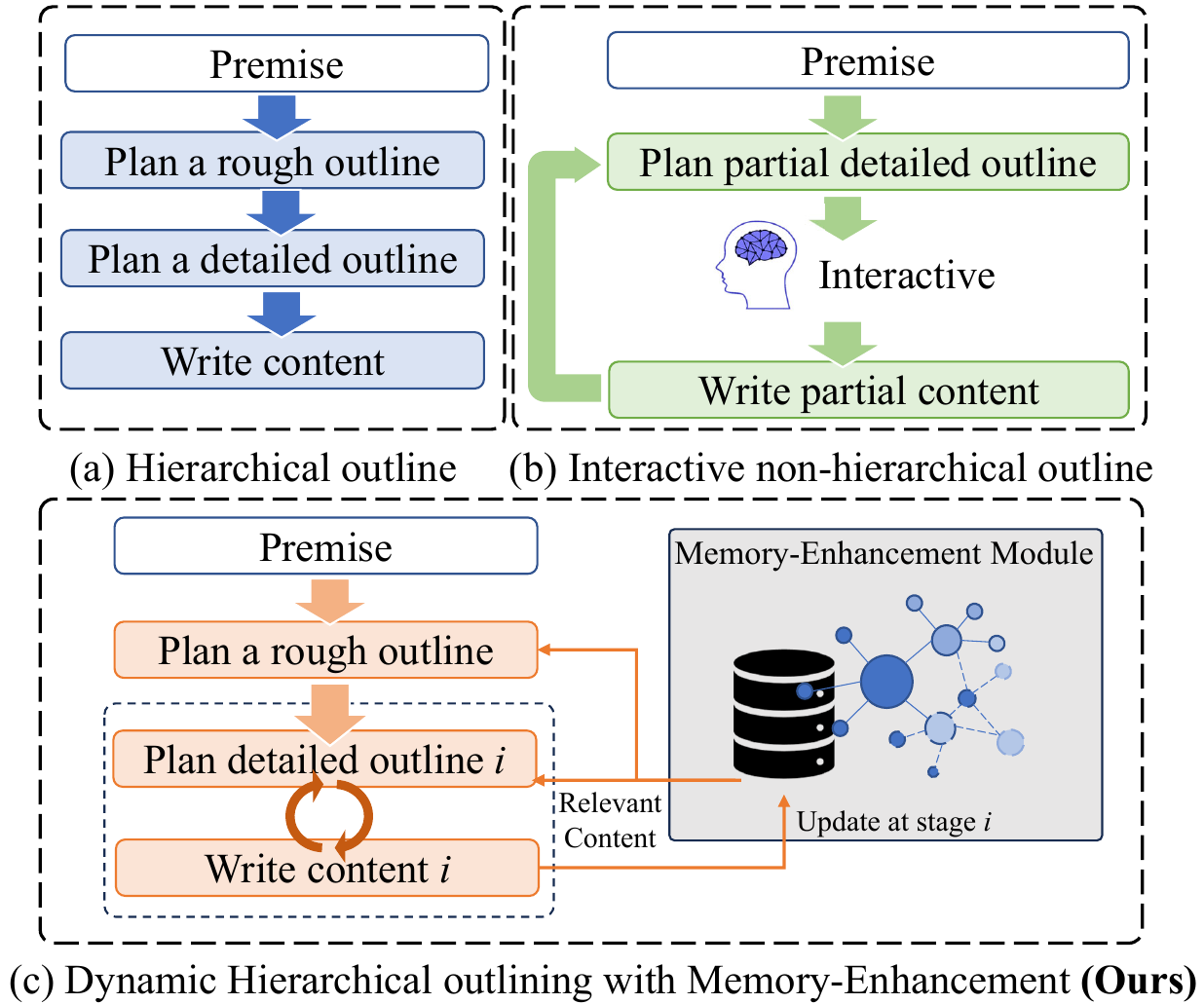}  
    \caption{Illustration and comparison of three strategies of long-form story generation. (a) applying a fixed hierarchical outline to guide the story generation, which is challenging to adapt to the uncertainty in story creation. (b) adapting to the uncertainty through interaction with humans, which allows for flexibility but lacks a high-level storyline to guide story development. Our method shown in (c) aims to enhance story coherence from both plot and expression and enjoys the advantages of the former two strategies to improve the coherence of the plot and reduce contextual conflicts.}  
    \label{method_compare}
    \vspace{-0.5cm}    
\end{figure} 

\hjw{Among natural language processing (NLP) tasks, the automatic generation of a long-form story is a representative task that requires creativity and long-term planning skills. By learning from human-written stories, an automated storyteller mimics humans and becomes competent in producing stories useful for various application scenarios, such as novel writing and interactive storytelling~\cite{riedl2010narrative}. Large Language Models (LLMs) are rapidly developing, generating a long-form story that further increases dramatically in length, complexity, and fluency~\cite{Re3}.}

Unfortunately, it is difficult for the LLMs to generate a long-form story maintaining contextual consistency in semantic and coherent plot development for the following reasons. 1) \textit{Memory limitations}: The black-box self-attention mechanism forms the core of LLMs for contextual connections and still suffers from the long-range dependency issue~\cite{vaswani2017attention}, making it hard for them to precisely and unambiguously recall and thereby leading to contextual incoherence. 2) \textit{Planning difficulties}: LLMs cannot effectively apply knowledge of coherent plot planning partly because their training involves learning from vast datasets that include a variety of texts, but they do not inherently understand or apply the principles of storytelling and thus hinders the generation of engaging stories with complete and fluent plots~\cite{xie-etal-2023-next}

\hjw{Existing long-form story generation methods often leverage higher-level attributes such as plots or commonsense knowledge~\cite{xie-etal-2023-next}, primarily aiming to enhance story development fluency, and can be divided into two kinds based on whether humans are involved. The first category emulates the human writing process by adopting a plan-and-write framework~\cite{Re3} in generating the long-form story (see Fig. \ref{method_compare}(a)). This approach separates the generation process into the planning and writing stages, utilizing detailed and plot-fluent outlines to guide the writing phase. While these methods can extend the length of the story and improve plot development fluency, the inflexibility of a fixed outline can impede its adaptability to uncertainty in the writing stage, often leading to plot incoherence, such as contextual repetition or conflicts.}Other works ~\cite{RecurrentGPT,brahman2020cue} explore generating content progressively without outlining based on human interaction and relevant preceding content \hjw{(see Fig. \ref{method_compare} (b))}. They can address the influence of local plot development caused by uncertainty in generation. However, the generated story lacks a macro-level of rational planning, affecting the plot completeness.

\hjw{To address the above limitations, we propose \textbf{D}ynamic Hierarchical \textbf{O}utlining with \textbf{M}emory-\textbf{E}nhancement long-form story generation method, called \textbf{DOME}, to generate the long-form story with coherent content and plot (see Fig. \ref{method_compare}(c)). \wqyr{\textbf{DOME} is based on the collaboration of a Dynamic Hierarchical Outline (DHO) mechanism and a Memory-Enhancement Module (MEM) to generate long-form story.}
Specifically, DHO mechanism to guide long-form story generation based on the plan-write writing framework~\cite{Re3} and the novel-writing theory \cite{TheoryFive}. The DHO mechanism requires the generation of a rough outline conforming to the novel writing theory to ensure plot completeness, and the dynamic planning of a detailed outline based on the rough outline during the writing process so that it can adapt to the uncertainty of the generation and improve the plot fluency. The MEM stores and accesses generated stories through temporal knowledge graphs and provides contextual content for outline planning and story writing to reduce content conflicts in long story texts. \wqyr{To automatically evaluate the contextual consistency, we propose a conflict detection matrix (named Temporal Conflict Analyzer) based on the information representation rules of the temporal knowledge graph and LLM.}}

\hjw{Our main contributions are as follows:}

\begin{itemize}
\vspace{-6pt}
\item 

\hjw{\textbf{A new paradigm for long-form story generation.} 
\wqyr{Since we find that existing methods for long-form story generation either cause plot incoherence due to fixed outlines or lack of fluency and readability due to poor macro-level planning, we propose a dynamic hierarchical outline (DHO) mechanism to fuse the planning and writing stages, making it adaptable to the generation uncertainty and improving plot coherence. Experiments show that DHO improves 6.87\% on ${Ent}$-2 metric.}}

\vspace{-6pt}
\item
\hjw{\textbf{A new approach to contextual conflict resolution.} We propose a Memory-enhancement module (MEM) based on the temporal knowledge graph to store and access the generated story. \wqyr{Applying LLM to perform semantic filtering on KG retrieval results keeps the conciseness of historical information and thus improves the contextual consistency of generated stories. 
}
Experiments show that MEM reduces conflicts by 87.61\%.}

\vspace{-6pt}
\item 
\textbf{A new evaluation for conflict detection.} To evaluate the degree of contextual consistency automatically, we propose a potential conflict detection method based on the information representation rules of the temporal knowledge graph and apply LLM to further determine whether a conflict exists. 
\wqyr{Experiments show that the judgment results of this method are consistent with human preferences.}

\end{itemize}

\begin{figure*}[t]

	\centering
	\includegraphics[width=0.96\textwidth]{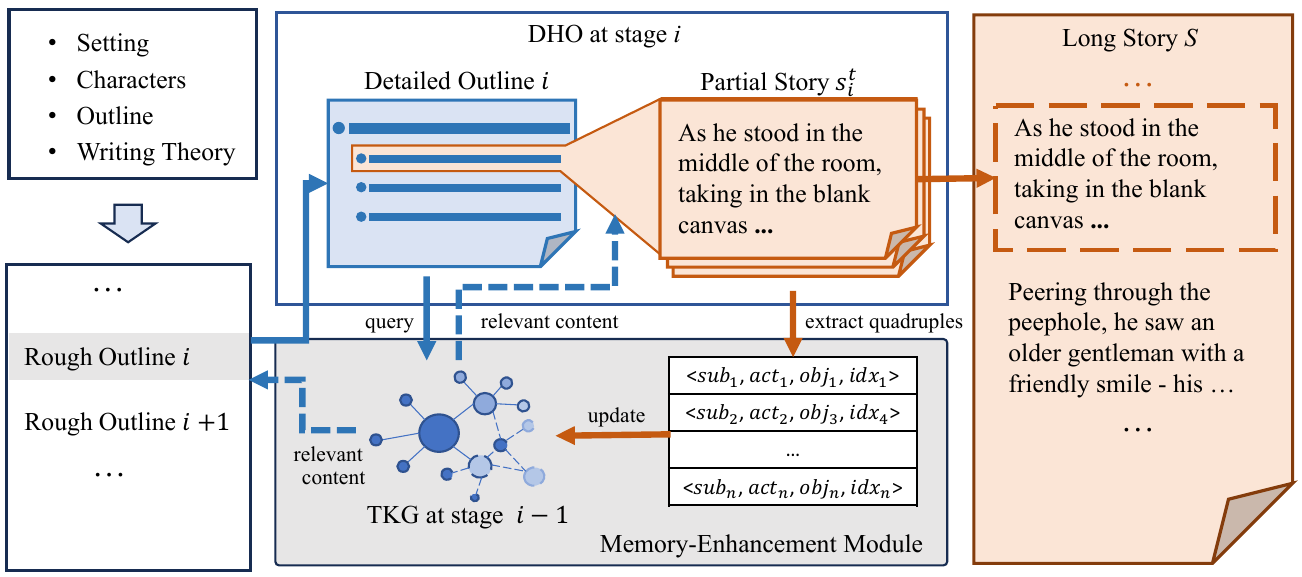}
	\caption{General diagram of proposed \methodname. The story generation process is divided into several stages based on the amount of rough outline.  At stage $i$, we expand rough outline $i$ into several detailed outlines based on the relevant content provided by MEM, and these outlines generate partial story sequentially based on their relevant content querying from MEM. Every generated partial story is stored in a temporal Knowledge graph for the following query.}
	\label{fig:our_method}
 \vspace{-6pt}
\end{figure*}

%% file: section/2_RelatedWork.tex
\section{Related Work}

\subsection{Long-form Story Generation}

Existing works have attempted to use language models to automatically generate novels \cite{Re3, DOC}. With the widespread use of LLMs, people begin to generate longer stories~\citep{RecurrentGPT}.Recent works for generating long-form stories can be classified into two categories based on whether human participation is engaged.The first category imitates the human writing process through a plan-and-write framework~\cite{planAndWrite}. ~\citet{facebook2018} proposes a hierarchical story generation method, which first generates a story premise and then generates the story based on it. \citet{planAndWrite} introduces a "plan-and-write" framework for open-domain story generation, which divides the writing process into planning and generating. \citet{Re3} further subdivides writing into plan, draft, rewrite, and edit module. \citet{DOC} makes efforts on outline control to generate a more detailed and hierarchical story. Although this technique can lengthen the narrative and enhance plot fluidity, the rigid nature of a predetermined outline can hinder flexibility in the writing phase, resulting in plot inconsistencies, such as repeated or contradictory contexts.

Other works \cite{RecurrentGPT, brahman2020cue} explore using human interaction to replace pre-generated outlines.~\citet{brahman2020cue} introduce a content-inducing approach, which involves human being using cue-phrases. \citet{RecurrentGPT} propose a language simulacrum of RNNs' recurrence, enabling human-guided story planning. This reduces plot inconsistencies but results in less fluent, readable narratives due to lack of microcontrol.

\subsection{KG-enhanced LLM Inference}

Knowledge Graphs (KGs) can effectively model the key information in text by triples in the form of $<'subject', 'action', 'object'>$~\citep{KGSurvey}. After the appearance of LLMs, like GPT-3.5~\cite{achiam2023gpt}, Llama 3~\cite{llama3} and Qwen1.5~\cite{bai2023qwen}, the integration of LLMs and KGs has attracted widespread attention due to the potential improvement KGs can make for LLMs~\cite{KGForLLMSurvey}. KG can be used as a knowledge base to provide references when LLM generates content. 
There are two kinds of works to inject the knowledge in KG into LLM. The first one fuse knowledge in KG and input by directly concatenating them in language level~\cite{KGForQA,wen2023mindmap,sun2021ernie} or in token level~\cite{zhang2019ernie,he2019integrating,su2021cokebert}. These methods can easily fuse the knowledge between them but with little interaction between knowledge and input. The second one applies the addition knowledge fusion module in which the knowledge is updated and simplified \cite{sun2021jointlk,zhang2022greaselm}.

%% file: section/Problem_definition.tex
\begin{figure*}[ht]
\centering
\includegraphics[width=0.95\linewidth]{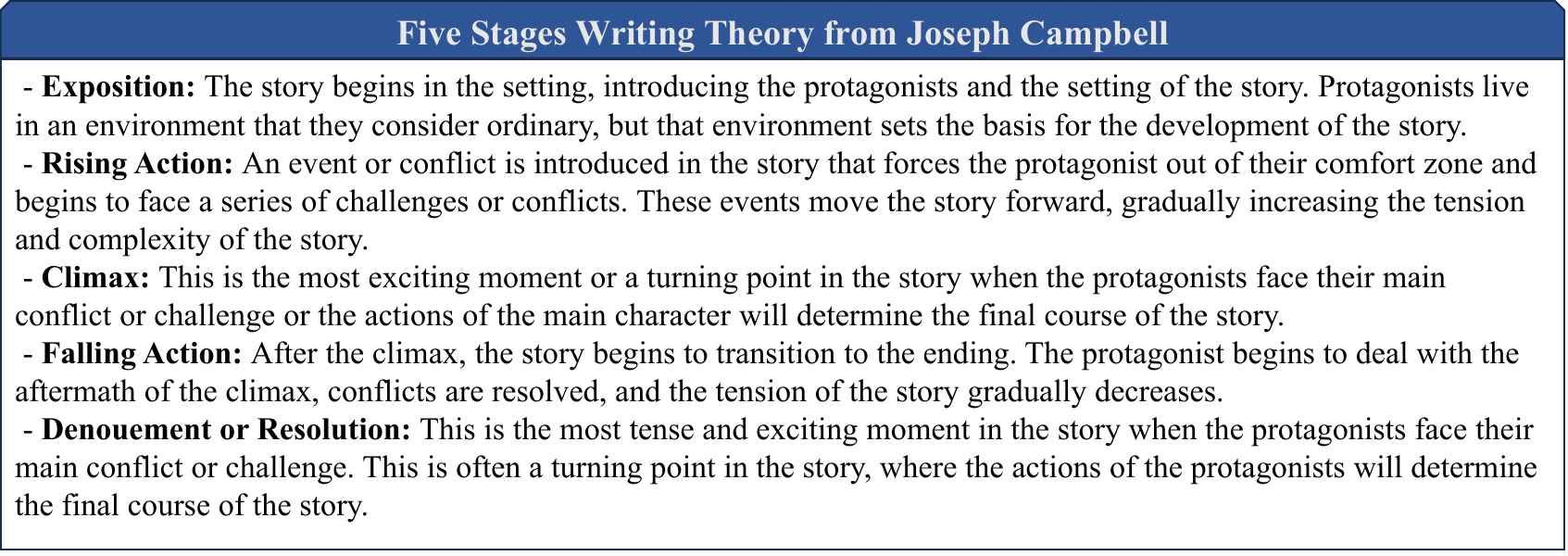}

    \caption{\hjw{Five stages novel writing theory from Joseph Campbell \cite{TheoryFive}.}}  
    \label{fig:novel}
\vspace{-6pt}
\end{figure*}

\section{Problem definition and motivations}

\subsection{Problem Definition}

\wqyr{For a given story premise ${I}$ from a writer, we design a framework ${F^*}$ to generate a long-form story ${S}$. The score of story $S$ on the plot coherence is $C^{Plot}$ and the score of story $S$ on the context coherence is $C^{Context}$. We aim to generate a story ${S}$ with improved $C^{Plot}$ and $C^{Context}$.}

\subsection{Motivaiton}

\wqyr{Existing long-form story generation methods separate planning and writing stages based on the plan-write framework, making it impossible to adapt to the uncertainty of the writing stage. Besides, stories generated through this two-stage method tend to stop abruptly at the beginning, resulting in missing plots. Another method includes human-involved detail outline preferences, improving the flexibility of outlines but making the overall storyline uncontrollable or incomplete. An intuitive idea arises: \textit{What if we could further improve story coherence from plot fluency and completion?}}

\wqyr{The answer is yes. We follow the idea of the hierarchical outline \cite{DOC} which is composed of the rough outline and the detailed outline. To ensure the plot completeness, we consider the rough outline as the macro plot guidance and apply the novel writing theory to guide its generation to ensure it contains all the story stages of the theory. The detailed outlines are expanded gradually based on the corresponding rough outline referred to the relevant content in the leading story to improve the adaptability to the generation uncertainty. 
In addition, with the increasing length of the generated story, the contextual inconsistency becomes obvious \cite{Re3}. Therefore, we design a memory module to store stories and access concise relevant content. Since KG can model information and retrieve the relevant information ~\cite{pan2024unifying}, we apply KG to store and access the semantically relevant content with a filter based on LLM. By providing accurate and concise relevant content during the generation stage, this module can improve contextual consistency.}

%% file: section/3_Method.tex
\begin{figure*}[t]
\centering
\includegraphics[width=0.96\linewidth]{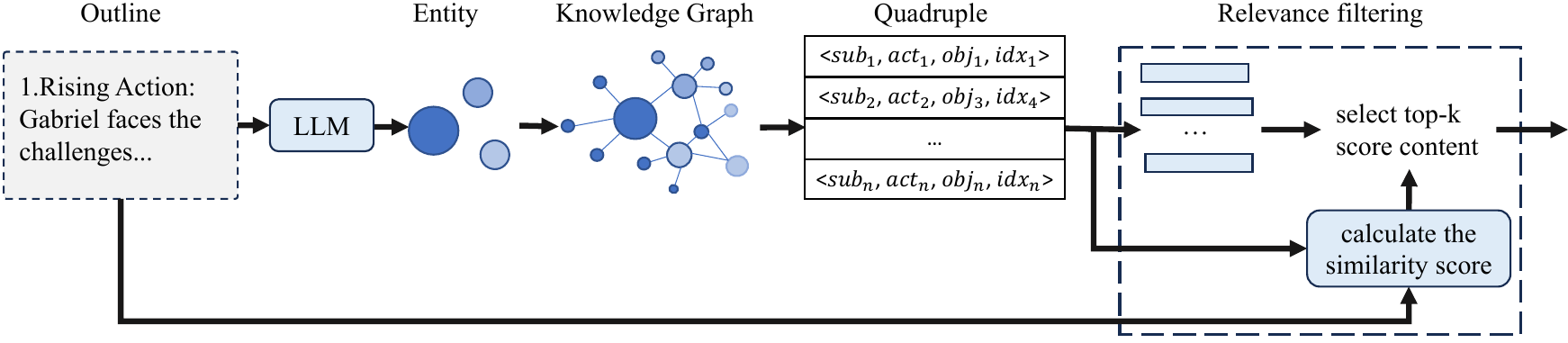}

    \caption{
    The details of querying for the relevant content. Notably the last arrow points to the historical information semantically related to the input content.
    }  
    \label{fig:query}
\vspace{-6pt}
\end{figure*}
\section{Proposed Method}
We propose a dynamic hierarchical outlining with memory-enhancement \hjw{long-form} story generation method, named \textbf{\methodname}, aiming to improve the story coherence from plot and context description. Based on the advanced understanding and generation ability of LLM \cite{liu2023pre}, \methodname~ is mainly composed of two parts: the dynamic hierarchical outline and writing module (\textbf{DHO}) and the KG-based memory-enhancement module (\textbf{MEM}). 
As shown in Fig.~\ref{fig:our_method}, given the user input $I$, including story setting, character introduction, and storyline requirements, the rough outlines of a story are planned at a time under the guidance of the writing theory, ensuring the completeness of the plot. Then, the detailed outline and the corresponding part of the story are generated alternately. It means the detailed outlines are not expanded by the rough outline incrementally until its previous stories are generated, enabling it to adjust to the uncertainty of the previously generated stories, and then the detailed outlines guide the generation of sequential stories. MEM stores user input at the very beginning and the generated story during the writing process. The module provides relevant content in natural language for the generation stage, ensuring contextual consistency. Through the collaboration of these two modules, \methodname~improve the story's coherence from plot and contextual consistency.

\subsection{\hjw{Dynamic Hierarchical Outline Mechanism}}
\label{sec:dho}

\hjw{The hierarchical outline  $H=\{R,D\}$ is composed of a rough outline $R$ and a detailed outline $D$.}\hjw{Previous works have used higher-level attributes like plots or commonsense knowledge to improve the quality of generated stories. We find that certain novel writing theories \cite{TheoryFive, TheorySeven, TheoryTwelve} enhance story generation. By incorporating the novel writing theory \cite{TheoryFive} (see Fig.~\ref{fig:novel}) into the rough outline planning stage, we improve the overall structure. This is done by stating the theory in the rough outline generation prompt. Thus, the rough outline $R$ was generated based on user input $I$:}
\begin{equation}
\label{eq:ROG}
R=LLM(WT,I,P_{rough\_utline}),
\end{equation}

\wqyr{
where $WT$ is the novel writing theory, $LLM(\cdot)$ is LLM inference operation, and $P_{rough\_outlie}$ is the prompt guiding LLM to generate a rough outline.}

The detailed outline \hjw{$D=\{d_i\}_{i=1}^{5}$} is generated step by step according to the corresponding part of the rough outline $r_i$ and its relevant content $RInfo._i$, ensuring its adaptation to the uncertainty in the story generation stage. 
\begin{equation}
\label{eq:DOG}
d_i=LLM(r_i,RInfo._i,P_{detailed\_outline}),
\end{equation}
\wqyr{
where $P_{detailed\_outline}$ is the prompt that indicates LLM to generate a detailed outline. The $d_i=\{do_i^t\}_{t=1}^{M}$ contains detailed outlines expanded from the rough outline. The $M$ is the number of detailed outlines expanded from the given rough outline $r_i$ which should be set at the beginning. 
}

The story content $s_i^t$ is generated step-by-step based on the detailed outline $do_i^t$ and relevant generated content $DInfo._i^t$ of the sub-detailed outline $do_i^t$, \hjw{which is as follows:} 
\begin{equation}
\label{eq:SG}
s_i^t=LLM(do_i^t,DInfo._i^t,P_{gen\_story}),
\end{equation}

\wqyr{where $P_{gen\_story}$ is the prompt that indicates LLM to generate a partial story. The detailed process of writing story $S$ in Algorithm \ref{alg:DO_S}. The details of prompts in the DHO are shown in Appendix~\ref{sup:prompt dho}.
}

\begin{algorithm}[t]
    \caption{Inference Pipeline for \methodname}
    \label{alg:DO_S}
    \begin{algorithmic}[1]
    \small
    \REQUIRE Novel writing theory $WT$, the large language model applied in DOME ${LLM}$, user input $I$, rough outline prompt $P_{rough\_outline}$, detailed outline prompt $P_{detailed\_outline}$, story generation prompt $P_{gen\_story}$.
    \STATE Initialize MEM $KG$.
    \STATE  Generate rough outline $R$ via Eqn. (\ref{eq:ROG}).
        \FOR{$r_i$ in $R=\{r_i\}_{i=1}^{i=5}$}
            \STATE Find relevant content $RInfo_i$ about $r_i$ from $KG$.
            \STATE Generate detailed outlines $d_i$ via Eqn. (\ref{eq:DOG}).
            \STATE Add $d_i$ into $D$.
            \FOR {$do_i^t$ in $d_i=\{do_i^t\}_{t=1}^{t=M}$}
                \STATE Find relevant content $DInfo._i^t$ about $do_i^t$ from $KG$.
                \STATE Generate partial story $s_i^t$ via Eqn. (\ref{eq:SG}).
                \STATE Add $s_i^t$ into story $S$.
                \STATE Store $s_i^t$ into $KG$.
            \ENDFOR
        \ENDFOR
        \RETURN $R$, $D$ and $S$.
    \end{algorithmic}
  
\end{algorithm}

\begin{figure*}[t]
\centering
\includegraphics[width=0.92\linewidth]{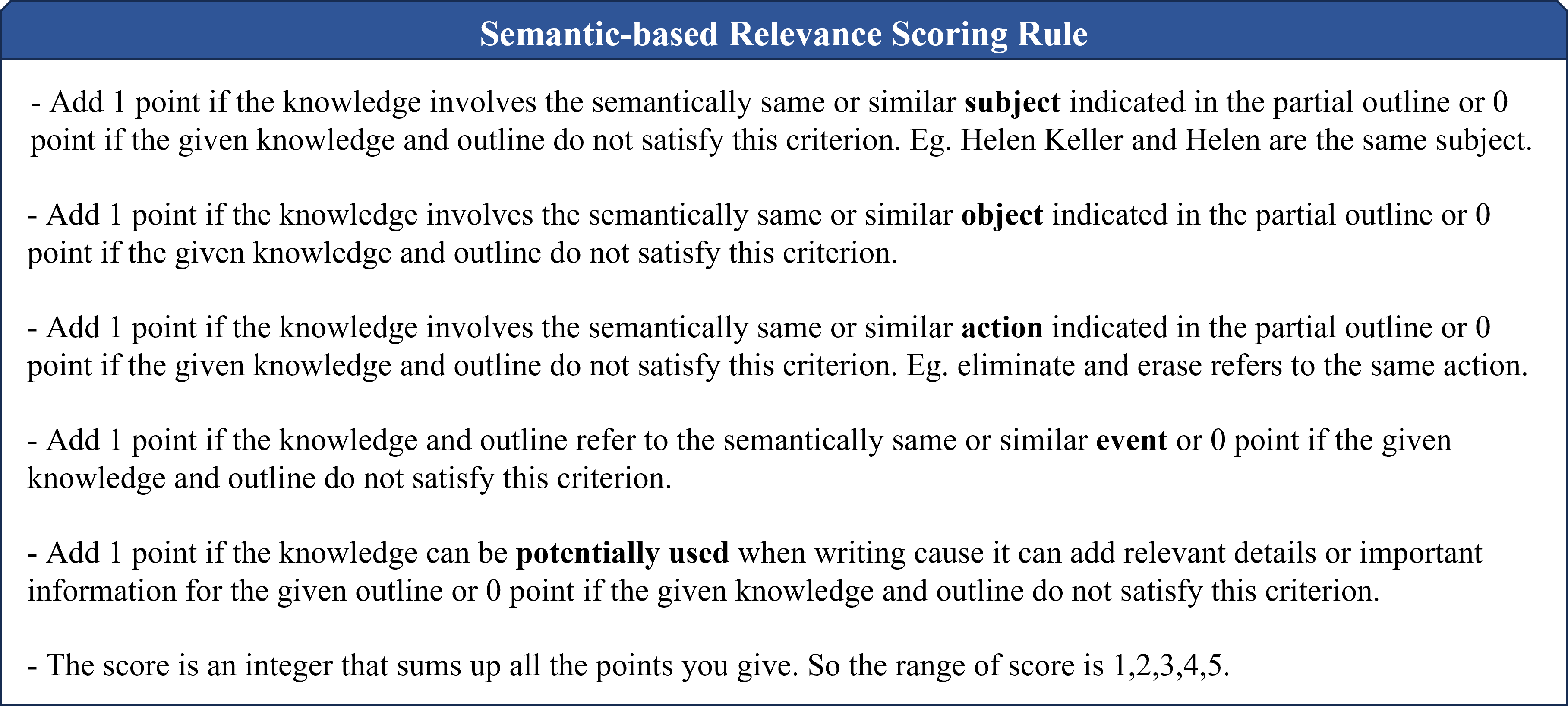}

    \caption{
    The criterion of filtering on semantic relevance.
    }  
    \label{fig:critition}
\vspace{-5pt}
\end{figure*}
\subsection{\hjw{Memory-Enhancement Module}}
\hjw{Recently, many works \cite{liu2024lostinmiddle} have} shown that LLMs tend to ignore the information in the middle of the long inputs. As the story generation process progresses, the increasing story length affects the LLM's attention on the relevant content when generating new content, increasing computing overhead \cite{choromanski2020rethinking} and even leading to contextual conflict \cite{Re3}. We propose an additional memory-enhancement module, named \textbf{MEM}, to store the generated story and provide concise relevant content.

KGs and their variants are well-established tools for content storage and query \cite{wen2023mindmap}. KGs extract the key information for the content like subject, action, and object and ignore unimportant information like adverbs and modifiers.
Besides, the structure of the knowledge graph supports reasoning and information fusion, which is conducive to information integration \cite{Pan2023UnifyingLL}.
Based on the analysis above, \hjw{the MEM} applies temporal knowledge graph (TKG) \cite{roddick2002survey} to store the generated story. TKG is stored by quadruples in the form of $<subject, action, object, index>$. The index means the chapter number of the information. The module stores the input story premise at the very beginning and the generated content and provides query-relevant content 
based on entity retrieval in TKG each time the DHO generates content.

\hjw{To store entity, we use LLM to extract triples for every sentence and then add the current chapter number to form quadruples.}
\wqyr{To access the relevant content (see Fig. \ref{fig:query}), we first conduct entity-based quadruple retrieval~\cite{reinanda2020knowledge} on TKG and then apply LLM to filter the retrieved results through semantics correlation by evaluating relevance based on rules (see Fig. \ref{fig:critition}). MEM can provide the $top$-$k$ most relevant information as query-relevant content, making it concise and semantically relevant with input (more details can be seen in Appendix~\ref{sup:mem prompt}).}

\subsection{\hjw{Temporal Conflict Analyzer}}

\wqyr{The evaluation of long-form stories in previous work mainly relies on humans, which is time-consuming and laborious \cite{DOC,improving-pacing}.}
MEM applies TKG to store story and it is easy for TKG to associate context \cite{wen2023mindmap}. Thus we propose an auto metric that measures the contextual consistency of the story by calculating the rate of conflict quadruples to total quadruples. To detect the conflict quadruples, the quadruples in TKG $Q=\{q_{i}\}_{i=1}^{N}$ of generated stories are grouped by rules (\hjw{see Fig.~\ref{fig:grules}  in Appendix \ref{sup:grules}}) sequentially and without repetition and the potential conflict information is aggregated. \wqyr{Since LLm-as-a-judge has been verified by experiments \cite{llm-as-a-judge} and it can evaluate some features of a text within a certain length according to rules \cite{achiam2023gpt, liu-etal-2021-durecdial}, we apply LLM to further detect the conflict quadruples based on time order. More details about the temporal conflict analyzer are reported in Appendix~\ref{sup:mem example}. }In this way, we can find the quadruples $Q^{conflict}=\{q_{i}^{conflict}\}_{i=1}^{m}$ containing conflict information. \hjw{The conflict rate is expressed as follows:}
\begin{equation}
\label{eq:cr.}
CR.=m/N \times 100\%,
\end{equation}
\hjw{where $N$ is the number of quadruples in TKG and $m$ is the number of conflict quadruples in TKG.}

%% file: section/4_Experiment.tex
\section{Experiment}
\label{experiment}

\begin{table*}
\small
    \centering
    \renewcommand{\tabcolsep}{3.0pt}
    \renewcommand{\arraystretch}{1.0}
    \begin{tabular}{ccccccccc}
    \hline
        \multirow{2}{*}{Methods} & \multicolumn{3}{c}{Auto Evaluation} & \multicolumn{5}{c}{Human Evaluation} \\
        \cmidrule[\cmidrulewidth](r{0.5em}){2-4}  \cmidrule[\cmidrulewidth](r{0.5em}){5-9}
        & Word Num. & $CR.\downarrow$  & ${Ent-2}\uparrow$ &$PCo.\downarrow$ &$PCoh.\downarrow$ &$Rel.\downarrow$ & $Int.\downarrow$ &$Ecoh.\downarrow$ \\
        
        \hline
        Llama3-70B-Instruct~\cite{llama3} &612.65  &6.78   &9.25 &3.77  &3.81  &3.58  &3.96  &3.72  \\
        Qwen1.5-72B-Chat~\cite{bai2023qwen} &495.70   &0.66  &9.06 &4.77  &4.80  &4.58  &4.96  &4.72   \\
        \hline
        Re$^3$~\cite{Re3} &3802.15 &0.77 &11.56 &2.73  &2.53  &2.96  &2.56  &2.83    \\
        DOC~\cite{DOC} &3904.90  &1.21 &11.55 &2.58  &2.43  &2.63  &2.34  &2.56  \\
        \rowcolor{mycyan}[1.2ex][1.5ex] \textbf{Ours (Qwen1.5-72B-Chat)} &\textbf{7113.75} &\textbf{0.56} &\textbf{12.29} &\textbf{1.15}  &\textbf{1.44}  &\textbf{1.26}  &\textbf{1.77} & \textbf{1.16}  \\
        \hline
    \end{tabular}
    \caption{Comparison with LLMs and SOTA baselines. We use the same premises as DOC~\citet{DOC}.}
    \label{tab:human_eval}
\end{table*}

\begin{table*}
\small
    \centering
    \renewcommand{\tabcolsep}{3.0pt}
    \renewcommand{\arraystretch}{1.0}
    \begin{tabular}{ccccccccc}
    \hline
        \multirow{2}{*}{Methods} & \multicolumn{3}{c}{Auto Evaluation} & \multicolumn{5}{c}{Human Evaluation} \\
        \cmidrule[\cmidrulewidth](r{0.5em}){2-4}  \cmidrule[\cmidrulewidth](r{0.5em}){5-9}
        & Word Num. & $CR.\downarrow$  & ${Ent-2}\uparrow$ &$PCo.\downarrow$ &$PCoh.\downarrow$ &$Rel.\downarrow$ & $Int.\downarrow$ &$Ecoh.\downarrow$ \\
        
        \hline
        w/o MEM  &6511.10  &4.52    &10.00 &1.88   &2.24  &1.97  &1.96   &2.22  \\
        w/o DHO &1471.90   &0.65  &11.50 &2.92  &2.36  &2.86  &2.88  &2.46   \\
        \rowcolor{mycyan}[1.2ex][1.5ex] \textbf{Ours (Qwen1.5-72B-Chat)} &\textbf{7113.75} &\textbf{0.56} &\textbf{12.29} &\textbf{1.20}  &\textbf{1.40}  &\textbf{1.17 }  &\textbf{1.62} & \textbf{1.32}  \\
        \hline
    \end{tabular}
    \caption{Ablation study of different modules.}
    \label{tab:Ablation}
\end{table*}

\textbf{Implementation details. }
\wqyr{\methodname~\footnote{The implementation are available at \url{https://github.com/Qianyue-Wang/DOME}}contains DHO and MEM. In DHO, every rough outline is expanded into 3 detailed outlines which means the $M$ mentioned in Section \ref{sec:dho} is set to 3. The embedding model we applied is bge-large-en-v1.5~\cite{chen2024bge}. In MEM, \wqyr{we apply cosine similarity~\cite{lahitani2016cosine} and set the filter threshold to 0.75 to implement the entity-based query.}}
Besides, we switch off the default historical content input of LLM where MEM is applied. The temperature of LLM is set to 0.5 and the max token of LLM is 1000 while other parameters are kept by default. All the experiments are conducted on 2$\times$A800 GPU with CUDA version 11.3.

\textbf{Datasets and baselines. }\wqyr{Following the settings of DOC~\cite{DOC}, we use its story premises\footnote{All the datasets are available at \url{https://github.com/Qianyue-Wang/DOME_dataset}} as the input and generate 20 long stories for evaluation. The details of the dataset are in Appendix~\ref{sup:Datasets Detsils}.}{We use Qwen1.5-72B-chat~\cite{bai2023qwen} for long-form story generation as it has considerable performance~\cite{chiang2024chatbot} and can be deployed locally. We compare \methodname~with two types of methods. The first is prompting LLM to generate stories as long as possible with input. The second are the state-of-the-art (SOTA) methods including DOC~\cite{DOC} and Re$^3$~\cite{Re3}. To validate the scalability of \methodname, we report the improvement When applying \methodname~ on Llama-3-70b-Instruct~\cite{llama3} and Yi1.5-34B-chat~\cite{young2024yi}.
 }

\textbf{Metrics. }
\wqyr{Since the content repetition is an important aspect reflecting the quality of stories in terms of coherence~\cite{Re3}, We apply the auto evaluation metrics including n-gram entropy~\cite{zhang2018generating} and set $n=2$ ($Ent$-2) to evaluate the degree of content repetition by measuring the diversity of vocabulary. Besides, we apply our proposed conflict rate ($CR.$) to evaluate the contextual consistency of the generated stores.} \par
\wqyr{To evaluate the alignment of the stories with human preferences in terms of contextual consistency and coherence and other basic story quality, we further conduct a human evaluation to evaluate the story quality of plot completeness, coherence, relevance, interest level, and expression coherence.} 
We compare all methods according to each indicator and calculate their average rank. The description of every metric is as follows: 
    1) Plot Completeness ($PCo.$): the extent to which the generated story covers all the stages mentioned in the story theory.
    2)  Plot Coherence ($PCoh.$): the fluency of the development of the generated story.
    3)  Relevance ($Rel.$): the consistency between the generated story and the input premise. \wqyr{It demonstrates the actual usability of the method.}
    4)  Interesting ($Int.$): the reading interest to the user. \wqyr{Since the reading interest can reflect peoples' preferences on the completeness and coherence of stories \cite{wang-etal-2023-improving-pacing}.}
    5)  Expression Coherence ($ECoh.$): the contextual consistency of the story.
More details of human devaluation are shown in Appendix~\ref{sup:human evaluate}.

\subsection{Comparison experiments}

\textbf{\methodname~is consistently better in auto evaluation.} From Table \ref{tab:human_eval}, the proposed \methodname~achieves superior performance in both $CR.$ and $Ent$-2. It also generates the longest content.
Specifically, \methodname~surpasses the LLM baseline, Qwen1.5-72B-Chat, by 35.7\% in $Ent$-2 and reduce the $CR.$ by 15.2\%. Additionally, \methodname~outperforms the Re$^3$ by 6.3\% in $Ent$-2 and reduces the $CR.$ by 27.3\%. The results indicate that our \methodname~delivers diverse content with fewer conflicts. We attribute the improvement to the following facts. Firstly, our TKG is capable of fine-grained modeling of generated stories. It provides accurate relevant content through semantic filtering during the generation stage, which ensures consistency and reduces contextual conflicts in the generated content. Besides, Our DHO dynamically adjusts the detailed outlines during the story generation process, thereby increasing the space for plot development and enhancing content diversity.

\textbf{\methodname~is consistently better in human evaluation.}We conduct a human evaluation to assess how well \methodname~aligns with human preferences. In Table \ref{tab:human_eval}, our \methodname~ranks first across all five metrics. The results demonstrate that MEM can provide relevant content during the generation stage, reducing the conflict due to the unclear memory of LLM itself~\cite{zhang2023siren} and thus ensuring contextual consistency and enhancing relevance and expression coherence. Furthermore, our DHO dynamically generates detailed outlines based on the development of the storyline, adapting to the uncertainty in generation and achieving improved fluency. Additionally, our novel-writing mechanism encourages large language models to create stories like writing a complete novel. This approach ensures the plot's completeness and to some extent, introduces fluctuations that capture and maintain reader interest.

\subsection{Ablation studies}

\textbf{Effectiveness of DHO.}\wqyr{We remove the DHO so that the outline is fixed and consistent with the input storyline.} As shown in Table \ref{tab:Ablation}, this variant exhibits higher content conflicts and reduced content \wqyr{variety}. Additionally, it ranks lowest across all human evaluation metrics. It supports the notion that dynamic adjustment of the detailed outline helps to control content generation, thereby ensuring plot coherence. Furthermore, the application of novel writing theory enhances plot completeness and thus improves the reading interest, \wqyr{this is because a complete story encourages the reader to read to the end.} See the example of DHO in Appendix~\ref{sup:dho example}.

\textbf{Effectiveness of MEM.}
Similarly, we remove the MEM to eliminate the knowledge graph for reference and resort to the multi-round chatting capability of LLM to generate content. Limited by computing cost, we set the maximum chat rounds to 2. \wqyr{As shown in Table \ref{tab:Ablation}}, the conflict rate significantly increases from 0.56 to 4.52 and $Ent$-2 reduces from 12.29 to 10.00 without MEM. Additionally, all human evaluation metrics deteriorate to varying degrees. These results indicate that MEM ensures contextual coherence by providing relevant content during generation, which also improves plot coherence and alignment with human preferences. See the example of MEM in Appendix~\ref{sup:mem example}.

\begin{figure}[t]
\centering
\includegraphics[width=1.0\linewidth]{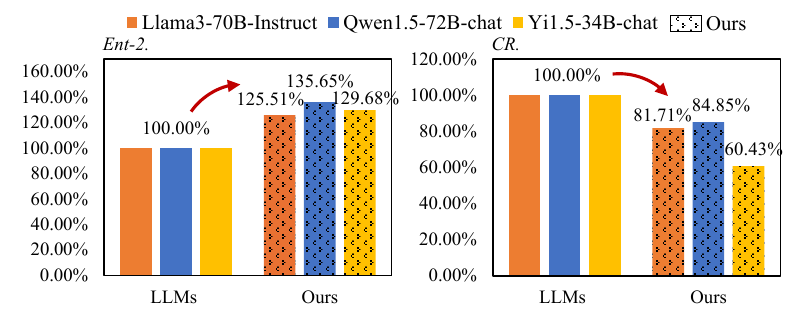}

    \caption{
    Scalability results of applying~\methodname~on different LLMs. Left: the increase rates of $Ent$-2 metric. Right: the decrease rates of $CR.$ metric.
    }  
    \label{fig:exp3}
\vspace{-6pt}
\end{figure}
\subsection{The Scalability of \methodname}

\wqyr{To demonstrate the applicability of \methodname~on different LLMs, we apply our method to  Yi1.5-34B-chat, Llama3-70B-Instruct and Qwen1.5-72B-chat. As illustrated in Fig.~\ref{fig:exp3}, \methodname~significantly reduces conflicts and enhances content diversity on all LLMs with different parameter scales. Specifically, \methodname~reduce the $CR.$ by more than 18\% on Llama3-70B-Instruct. It also improves the $Ent$-2 by more than 29\% on Yi1.5-34B-chat. This scalability is due to the straightforward and easy-to-follow prompts in our \methodname, making it easy to be adaptable across different LLMs. }

\begin{table}[t]
\small
    \centering
    \renewcommand{\tabcolsep}{1pt}
    \renewcommand{\arraystretch}{1.2}
    \begin{tabular}{ccccccccc}
    \hline
        LLM (w \methodname) & Nodes	&Relations	&Quadruples &API calls\\ \hline
        Yi1.5-34B-chat & 691.3	&328.45 &354.55&16\\ 
        Llama3-70B-instruct& 844.6	& 345.65& 434.0&16\\
        Qwen1.5-72B-chat & 791.35	&258.23	&404.85 &16\\
        \hline
    \end{tabular}
    \caption{\wqyr{The computation and storage cost of \methodname. Notable the API calls means the average number of API calls for KG construction.}}
    \label{tab: storage cost}
    \vspace{-6pt}
\end{table}

\subsection{\wqyr{Computation and storage cost of \methodname}}
\wqyr{We report the computation and storage cost for constructing the knowledge graph dynamically, as shown in Table \ref{tab: storage cost}. With acceptable additional computation and storage overhead, the proposed method improves the quality of generating long-form stories. Specifically, as shown in Tables \ref{tab:human_eval} and \ref{tab: storage cost}, the proposed DOME only adds 791 nodes of storage overhead and 16 LLM API calls to construct the KG dynamically, but it outperforms the SOTA methods in all performance metrics. This is because the structured storage of KG and the access advantages of KG facilitate LLM in achieving accurate and concise relevant memory and reducing context inconsistency.}

%% file: section/5_Analysis.tex
\section{Conclusion}
\hjw{In this paper, we propose a dynamic hierarchical outlining with memory-enhancement, named \textbf{\methodname}, to generate a coherent long-form story. 
Specifically, we propose a dynamic hierarchical outline (DHO) mechanism for long-form story generation, based on the plan-write framework and novel-writing theory. The DHO mechanism creates a rough outline that aligns with novel-writing theory and dynamically generates it during the writing process, enhancing plot fluency. Additionally, the Memory-Enhancement Module (MEM) uses temporal knowledge graphs to store and access generated stories, providing contextual content for both detailed outline planning and story writing, and reducing contextual conflicts. Lastly, the temporal conflict analyzer detects potential conflicts using temporal knowledge graphs and integrates with LLMs to automatically assess the contextual consistency of the generated text. 
Experiments demonstrate that \methodname~significantly improves the coherence and overall quality of generated long-form stories from plot and expression compared to SOTA methods.
}

%% file: section/6_Limitation.tex
\section*{Limitations} 
The lack of automatic evaluation matrics constrains our experiments. Specifically, we have to rely on human evaluation to evaluate the quality of the generated stories which is time-consuming and costly. Besides, the amount of experimental data is limited since there are no specific datasets for long-form story generation and we follow the experiment setting of \cite{DOC}. In addition, our framework requires massive LLM API calls which are time-consuming. On average, Generating a story requires about 200 API calls. Massive API calls result in a limited story generation speed, taking about 4 hours to complete a 7,000-word story. Thus the long text generation and a large number of API calls limit our usage of paid closed-source LLMs, such as ChatGPT.
Although we report the result about the extensibility of our framework, many steps in our framework are realized based on custom-designed prompts and it is better to re-design these prompts to achieve better performance.

\wqyr{Although experiments demonstrate the effectiveness and scalability of our proposed \methodname, the prompts in \methodname~are designed based on human experience. Therefore, the use of automatic prompts may further exploit the capabilities of \methodname. However, automatic prompting methods often struggle with capturing the nuanced context required for a specific task \cite{si2023prompting}. Applying them to long-form story generation tasks may not ensure the completeness of long-form stories and further affects the effectiveness and scalability of \methodname.}

%% file: section/Appendix.tex
\appendix

\section{Datasets Detsils}
\label{sup:Datasets Detsils}
We follow the setting of DOC \cite{DOC} which uses 20 story premises generated by InstructGPT3-175B as input for story generation. Every premise includes a story setting, a character introduction, and a necessary storyline. An example of an input is shown in the Fig.~\ref{tab:input premise}

\begin{figure*}[htbp]
\centering
\begin{tcolorbox}
        \textbf{Setting}\\
        The story is set in the inner city of a large metropolitan area.\\
        \\
        \textbf{Character Introdution}\\
        Gary Saunders: Gary Saunders is a teenage boy who lives in the inner city.\\
        Shannon Doyle: Shannon Doyle is a young woman in her early twenties.\\
        Mike Doyle: Mike Doyle is Shannon's father and a successful journalist.\\
        Lena Saunders: Lena Saunders is Gary's mother and a local business owner.\\
        \\
        \textbf{Necessary Storyline}\\
        1. Shannon's father, Mike, dies unexpectedly, leaving her determined to follow in his footsteps \\~~~~and become a successful journalist.\\
        2. Shannon lands her first major assignment, a feature on the inner city, but quickly discovers \\~~~~that the ugly reality of life in the city is far different from the dream she imagined.\\
        3. With the help of her new friend, Gary, Shannon comes to understand the harsh realities of life in the inner city and learns \\~~~~that sometimes the truth is much more than just a story.\\
\end{tcolorbox}
\caption{The example of input.}
\label{tab:input premise}
\end{figure*}


\section{Prompt Details}
\label{sep:prompt}
We design prompts for LLM to finish specific tasks in our framework and evaluate the contextual consistency in ${CR.}$. 

We design our prompt following the suggestions instead of applying the automatic prompt method since the automated prompt methods often struggle with capturing the nuanced context required for specific tasks \cite{si2023prompting} and may affect the generation quality.

\subsection{Prompts in DHO}
\label{sup:prompt dho}
In \textbf{DHO}, we prompt(shown in Fig.~\ref{tab:rough outline prompt}) LLM to plan a rough outline based on input and referring to the story writing theory. The detailed outline is planned by prompting(shown in Fig.~\ref{tab:detailed outline prompt}) LLM to realize based on the corresponding part of the rough outline referring to the relevant content.\par

\begin{figure*}[htbp]
\centering
\begin{tcolorbox}
        Based on the given novel story writing theory delimited by, the given novel setting delimited by, the given character introduction delimited by, and the given novel outline delimited by, plan the storyline of an episodic lone story.\\
The storyline contains the same number of parts as the theory.\\
The storyline you plan needs to be labeled concerning the description of each chapter according to the storyline planning theory.\\
\\
story writing theory:\{theory\}\\
Novel setting:\{setting\}\\
Character introduction:\{character\}\\
The general story:\{outline\}\\
\\
The output should be a markdown code snippet formatted in the following schema, including the leading and trailing "```json" and "```":\\
\\
Separate the five json objects with commas.\\
Contain all the json objects into a list object.\\
Nothing but your storyline should be contained in the answer.\\
Your storyline:\\
\end{tcolorbox}
\caption{The prompt for plan a rough outline.}
\label{tab:rough outline prompt}
\end{figure*}

\begin{figure*}[htbp]
\centering
\begin{tcolorbox}
       Your task is to generate chapter outlines based on the given volume of the novel. \\
    Below are some steps to help you complete the task:\\
    
    1. Determine how many chapters the given volume can be expanded into. You can determine the amount of chapters based on the stage of the volume in the story. The more intermediate a stage is, the more it needs to be expanded into more chapters, but no more than 5 normally each volume expands into 2 or 3 chapters. .\\
    2. Based on the content of the volume, determine the content for an outline of each chapter. The outline between chapters is not repeated.\\
    3. The generated chapter outline, should refer to the relevant historical information and the previous chapter outline to make the outline consistent with the previous text in description and plot development.\\
    \\
    \#\# Input\\
    the current volume outline: \{volume outline\}\\
    the stafe of the volume: \{stage\}\\
    the outline of the previous chapter: \{last chapter\}\\
    the related background: \{history\}\\
    \\
    \#\# Output Format \\
    Please follow the output format strictly to ensure the consistency of the generated detailed chapter outline.\\
    Nothing but only the three(3) chapter outlines should be included in the output.\\
    Following the format below:\\
    - Chapter Outline 1: \\
    - Chapter Outline 2: \\
    - Chapter Outline 3: \\
   \\
    Your result:\\
\end{tcolorbox}
\caption{The prompt for plan a detailed outline.}
\label{tab:detailed outline prompt}
\end{figure*}

\subsection{Prompts in MEM}
\label{sup:mem prompt}
In \textbf{MEM}, quadruples are extracted from the generated story based on the few-shot learning ability of LLM, and the prompt is shown in Fig.~\ref{tab:extract quadruple}. When querying relevant content from TKG, the entities from subjects and objects of quadruples extracted from a query are used to retrieve quadruples in TKG. LLM evaluates the relevant scores between the query and the retrieved quadruples based on the prompt shown in Fig.~\ref{tab:relevance}.

\begin{figure*}[htbp]
\centering
\begin{tcolorbox}
       Your task is summarizing in the form of triples according to the given text.\\
\\
    There is an example:\\
    given text:\\
    Lily lived in a quaint rural town, surrounded by lush greenery and rolling hills. Despite the tranquility of her surroundings, she often felt restless and yearned for adventure. One day, while exploring the dense jungle that lay beyond her town, Lily stumbled upon an ancient diary in her grandmother's attic. The pages were yellowed with age, and the ink had faded over time, but Lily could still make out the words written within.\\
    output triples:\\
    1.(Lily, lives in, quaint rural town)\\
    2. (quaint rural town, characterized by, lush greenery)\\
    3. (quaint rural town, characterized by, rolling hills)\\
    4. (Lily, feels, restlessness)\\
    5. (Lily, yearns for, adventure)\\
    6. (dense jungle, located beyond, quaint rural town)\\
    7. (Lily, discovers, ancient diary)\\
    8. (ancient diary, found in, grandmother's attic)\\
    9. (ancient diary, characterized by, yellowed pages)\\
    10. (ancient diary, characterized by, faded ink)\\
\\
    For each sentence delimited by '.' in the given text, you should follow these steps to extract the triples:\\
    step1:Find all the verbs without any modifiers in the sentence.\\
    step2:Find the subject and object without any modifiers for each verb from the step 1.\\
    If the subject or object of the verb is a pronoun like he, she, it, there, that, they, or those, replacing them with what they refer to by context. \\
    If you still can not determine what the pronoun refers to, use someone to replace it.\\
    Make sure every element in triple is clear when understood alone, that means there is no pronoun in triples.\\
    For a subject or object modified by a clause, treat the clause as a new sentence and follow the previous steps to extract a new triple for it.\\
    step3:Normalize the above results in triples which are in the form of (subject, verb, object) for each verb and number them by 1.,2.,3. and so on.\\
    Make sure there are only 3 elements in each triple. More or less elements are not acceptable.\\
    Make sure the triples are in the correct form which is the same as the given output example and the elements are clear and understandable.\\
\\
    The given text:\{text\}\\
    Your result:\\
\end{tcolorbox}
\caption{The prompt for extract quadruple from story.}
\label{tab:extract quadruple}
\end{figure*}

\begin{figure*}[htbp]
\centering
\begin{tcolorbox}
Scoring the Degree of correlation between the given outline and the given knowledge using the scoring criteria described below.\\
Points are accumulated based on the satisfaction of each criterion:\\
\\ 
    - Add 1 point if the knowledge involves the semantically same or similar subject indicated in the partial outline or add 0 point if the given knowledge and outline do not satisfy this criterion. Attention : Helen Keller and Helen refer to the same subject\\
    - Add 1 point if the knowledge involves the semantically same or similar object indicated in the partial outline or add 0 point if the given knowledge and outline do not satisfy this criterion.\\
    - Add 1 point if the knowledge involves the semantically same or similar action indicated in the partial outline or add 0 point if the given knowledge and outline do not satisfy this criterion.Attention : eliminate and erase refers to the same action\\
    - Add 1 point if the knowledge and outline refer to the semantically same or similar event or add 0 point if the given knowledge and outline do not satisfy this criterion.\\
    - Add 1 point if the knowledge can be potentially used when writing cause it can add relevant details or important information for the given outline or add 0 point if the given knowledge and outline do not satisfy this criterion.\\
    - The score is an integer that sums up all the points you give. So the range of scores is 1,2,3,4,5. \\
\\
The given outline:\{outline\}\\
the given knowledge:\{triplesentence\}\\
\\
Output Format\\
There are 3 parts in your generated output.\\
Please strictly follow the format to ensure the correct evaluation of the relevance.\\
Nothing but the following parts you make should be included in the output.\\
Part1 Score Results and their Reasons:\\
for criterion 1. My result is: add (0 or 1).Because:....\\
for criterion 2. My result is: add (0 or 1).Because:....\\
for criterion 3. My result is: add (0 or 1).Because:....\\
for criterion 4. My result is: add (0 or 1).Because:....\\
for criterion 5. My result is: add (0 or 1).Because:....\\
Part2 Sum Up:\\
Summing up all the score results for each criterion:\\
eg.1+1+1+0+0=3\\
Part3 total score\\
Score: \\
\\
Follow all the information above to generate the formatted output.\\
Do not contain any additional information except the output parts.\\
Your Output:\\
\end{tcolorbox}
\caption{The prompt for score relevance between the description and the given outline.}
\label{tab:relevance}
\end{figure*}

\subsection{Prompts in Temporal Conflict Analyzer}
In the temporal conflict analyzer, we first apply rules to group quadruples and then use LLM to further determine the existence of contextual conflict. In detail, quadruples of every group are first described in natural language and then judged to be reasonable or not based on common sense. All these are completed by prompting LLM applied in the framework. Evaluating the contextual consistency by LLM is also based on prompts and is further divided into two steps: 1)describe the information expressed by a group of quadruples in natural language based on the grouping feature(shown in Fig.~\ref{tab:group trans1}, Fig.~\ref{tab:group trans2}, Fig.~\ref{tab:group trans3}, Fig.~\ref{tab:group trans4} and Fig.~\ref{tab:group trans5}.)
2)judge if the description is reasonable based on the grouping feature(shown in Fig.~\ref{tab:judge1}, Fig.~\ref{tab:judge2}, Fig.~\ref{tab:judge3}, Fig.~\ref{tab:judge4}, Fig.~\ref{tab:judge5}). The mining on the group from filtered information use the same information expression prompts which are shown Fig.~\ref{tab:group trans1}, Fig.~\ref{tab:group trans2}, Fig.~\ref{tab:group trans3}, Fig.~\ref{tab:group trans4}.

\begin{figure*}[htbp]
\centering
\begin{tcolorbox}
Generate a logical summary from the given input list. \\
The input list contains a series of quadruples, each containing the following elements:\\
The first element is the entity making the action.\\
The second element is the action.\\
The third element is the giving object of the action.\\
The fourth element is a number that represents the chapter in which the action took place.\\
\\
The first two elements of the given series quadruple are the same, indicating that the sender of the action and the action are always the same. You should summarize the information from this perspective.\\
\\
Here is a reference example:\\
Sample input: [("Bob", "hit", "Jane", 1), ("Bob", "hit", "Lily", 1), ("Bob", "hit", "Mary", 2)]\\
Example output: Bob hit Jane and Lily in Chapter 1 and then hit Mary in chapter 2.
Xiao Ming typed small green and small blue in the first chapter, and small red in the second chapter.\\
\\
The output should express the information smoothly and match the accuracy of the input.\\
The output should be expressed in just one sentence.\\
\\
The list given:\{inlis\}\\
Your results:\\
\end{tcolorbox}
\caption{The prompt for describing the information based on quadruples grouped by rule 1.}
\label{tab:group trans1}
\end{figure*}

\begin{figure*}[htbp]
\centering
\begin{tcolorbox}
Generate a logical summary from the given input list. \\
The input list contains a series of quadruples, each containing the following elements:\\
The first element is the entity making the action.\\
The second element is the action.\\
The third element is the giving object of the action.\\
The fourth element is a number that represents the chapter in which the action took place.\\
\\
The first three elements of the given series quadruple are the same, indicating that the occurrence of actions remains the same in each chapter. You should summarize the information from this perspective.\\
\\
Here are some reference examples:\\
Sample input: [("Bob", "hit", "Jane", 1), ("Bob", "hit", "Jane", 2), ("Bob", "hit", "Jane", 3)]\\
Example output: Bob hit Jane from chapter 1 to chapter 3.\\
\\
Sample input: [("Bob", "hit", "Jane", 1), ("Bob", "hit", "Jane", 3)]\\
Example output: Bob hit Jane from chapter 1 and chapter 3.\\
\\
The output should express the information smoothly and match the accuracy of the input.\\
The output should be expressed in just one sentence.\\
\\
The list given:\{inlist\}\\
Your results:\\
\end{tcolorbox}
\caption{The prompt for describing the information based on quadruples grouped by rule 2.}
\label{tab:group trans2}
\end{figure*}

\begin{figure*}[htbp]
\centering
\begin{tcolorbox}
Generate a logical summary from the given input list. \\
The input list contains a series of quadruples, each containing the following elements:\\
The first element is the entity making the action.\\
The second element is the action.\\
The third element is the giving object of the action.\\
The fourth element is a number that represents the chapter in which the action took place.\\
\\
The second and third elements of the given quadruples are the same, indicating that the receiver of the action and the action are always the same. You should summarize the information from this perspective.\\
Here is an example:\\
input: [("Lily", "hit", "Jane", 1), ("Bob", "hit", "Jane", 1), ("Emma", "hit", "Jane", 3)]\\
Example output: Jane was hit by Liy and Bob in chapter 1 and she was hit by Emma in chapter 3.\\
\\
The list given:\{inlist\}\\
The output should express the information smoothly and match the accuracy of the input.\\
The output should be expressed in just one sentence.\\
Your results:\\
\end{tcolorbox}
\caption{The prompt for describing the information based on quadruples grouped by rule 3.}
\label{tab:group trans3}
\end{figure*}

\begin{figure*}[htbp]
\centering
\begin{tcolorbox}
Generate a logical summary from the given input list. \\
The input list contains a series of quadruples, each containing the following elements:\\
The first element is the entity making the action.\\
The second element is the action.\\
The third element is the giving object of the action.\\
The fourth element is a number that represents the chapter in which the action took place.\\
\\
The first and third elements of the given quadruples are the same, indicating that the sender of some action and the receiver of some action are always the same.You should summary the information from this perspective.\\
There are some example:\\
input: [("Lily", "hate", "Jane", 2), ("Lily", "Love", "Jane", 1)]\\
Example output: Lily loves Jane at chapter 1 but grows to hate Jane at chapter 2.\\
input: [("Lily", "kill", "Jane", 2), ("Lily", "hate", "Jane", 1)]\\
Example output: Lily hates Jane at chapter 1 and kills Jane at chapter 2.\\
\\
The list given:\{inlist\}\\
The output should express the information smoothly and match the accuracy of the input.\\
The output should be expressed in just one sentence.\\
Your results:\\
\end{tcolorbox}
\caption{The prompt for describing the information based on quadruples grouped by rule 4.}
\label{tab:group trans4}
\end{figure*}

\begin{figure*}[htbp]
\centering
\begin{tcolorbox}
Generate a logical summary from the given input list. \\
The input list contains a series of entities including their attributes and the chapter numbers in which they appear.\\
Here is a summary from a list of grouped quadruples:\\
input:[['small garden',1],['unfinished garden',3],['beautiful garden',2]]\\
Example output: the garden is small in chapter 1 and beautiful in chapter 2 but in chapter 3 ,it is unfinished.\\
\\
The list given:\{inlist\}\\
The output should express the information smoothly and match the accuracy of the input.\\
The output should be expressed in just one sentence.\\
Your results:\\
\end{tcolorbox}
\caption{The prompt for describing the information based on quadruples grouped by rule 5.}
\label{tab:group trans5}
\end{figure*}

\begin{figure*}[htbp]
\centering
\begin{tcolorbox}
Determine if the description given is reasonable.
\\
The sender of the action and the main part of the action are always the same.\\
You need to determine whether the recivers of the action are reasonable as the knowledge described over Chapter identification.\\
You need to judge based on context of knowledge and Combined with the modified part of recivers to make the judgment.\\
The input:\{description\}\\
The output should be a markdown code snippet formatted and the format is : 
\{
        "result": string  Was a char chosen from 'Y' or 'N' that describes the judgment result. If the judgment result is 'Y', it means that the two schemas have conflicts. If the judgment result is 'N', it means that the two schemas do not have conflicts.
        "explanation": string Was a string that describes how you judge the conflict or conflict-free between the two schemas.
\}
Please be strict in your judgment and consider the chronological order of the attributes.\\
Strictly judge based on the given information, do not add any information to make some unreasonable science into reasonable.\\
Your results:\\
\end{tcolorbox}
\caption{The prompt for judging whether the description is reasonable based on rule 1.}
\label{tab:judge1}
\end{figure*}

\begin{figure*}[htbp]
\centering
\begin{tcolorbox}
Determine if the description given is reasonable.
\\
The sender,the reciver of the action and the main part of the action itself are always the same but it happend at different chapter.\\
You need to determine whether the information in the knowledge can be maintained over Chapter identification in the knowledge\\
You need to judge based on context of knowledge and your basic Semantic knowledge.\\
The input:\{description\}\\
The output should be a markdown code snippet formatted and the format is : 
\{
        "result": string  Was a char chosen from 'Y' or 'N' that describes the judgment result. If the judgment result is 'Y', it means that the two schemas have conflicts. If the judgment result is 'N', it means that the two schemas do not have conflicts.
        "explanation": string Was a string that describes how you judge the conflict or conflict-free between the two schemas.
\}
Please be strict in your judgment and consider the chronological order of the attributes.\\
Strictly judge based on the given information, do not add any information to make some unreasonable science into reasonable.\\
Your results:\\
\end{tcolorbox}
\caption{The prompt for judging whether the description is reasonable based on rule 2.}
\label{tab:judge2}
\end{figure*}

\begin{figure*}[htbp]
\centering
\begin{tcolorbox}
Determine if the description given is reasonable.
\\
The reciver of the action and the main part of the action itself are always the same but the sender of the actition is different.\\
You need to determine whether the senders of action are reasonable as the knowledge described over Chapter identification.\\
You need to judge based on context of knowledge and Combined with the modified part of senders to make the judgment.\\
The input:\{description\}\\
The output should be a markdown code snippet formatted and the format is : 
\{
        "result": string  Was a char chosen from 'Y' or 'N' that describes the judgment result. If the judgment result is 'Y', it means that the two schemas have conflicts. If the judgment result is 'N', it means that the two schemas do not have conflicts.
        "explanation": string Was a string that describes how you judge the conflict or conflict-free between the two schemas.
\}
Please be strict in your judgment and consider the chronological order of the attributes.\\
Strictly judge based on the given information, do not add any information to make some unreasonable science into reasonable.\\
Your results:\\
\end{tcolorbox}
\caption{The prompt for judging whether the description is reasonable based on rule 3.}
\label{tab:judge3}
\end{figure*}

\begin{figure*}[htbp]
\centering
\begin{tcolorbox}
Determine if the description given is reasonable.
\\
The sender and the receiver of the action are always the same but the action may be different.\\
You need to determine whether the actions can co-exist as what the knowledge described over Chapter identification.\\
You need to judge based on the context of knowledge and Combine with the modified part of the action to make the judgment.\\
The input:\{description\}\\
The output should be a markdown code snippet formatted and the format is : 
\{
        "result": string  Was a char chosen from 'Y' or 'N' that describes the judgment result. If the judgment result is 'Y', it means that the two schemas have conflicts. If the judgment result is 'N', it means that the two schemas do not have conflicts.
        "explanation": string Was a string that describes how you judge the conflict or conflict-free between the two schemas.
\}
Please be strict in your judgment and consider the chronological order of the attributes.\\
Strictly judge based on the given information, do not add any information to make some unreasonable science into reasonable.\\
Your results:\\
\end{tcolorbox}
\caption{The prompt for judging whether the description is reasonable based on rule 4.}
\label{tab:judge4}
\end{figure*}

\begin{figure*}[htbp]
\centering
\begin{tcolorbox}
Determine if the description given is reasonable.
\\
The description given is the change in the state of the same entity over time.\\
The input:\{description\}\\
The output should be a markdown code snippet formatted and the format is : 
\{
        "result": string  Was a char chosen from 'Y' or 'N' that describes the judgment result. If the judgment result is 'Y', it means that the two schemas have conflicts. If the judgment result is 'N', it means that the two schemas do not have conflicts.
        "explanation": string Was a string that describes how you judge the conflict or conflict-free between the two schemas.
\}
Please be strict in your judgment and consider the chronological order of the attributes.\\
Strictly judge based on the given information, do not add any information to make some unreasonable science into reasonable.\\
Your results:\\
\end{tcolorbox}
\caption{The prompt for judging whether the description is reasonable based on rule 5.}
\label{tab:judge5}
\end{figure*}

\subsection{Prompts in experiments}
\label{sup:llm prompt}
One of our compared methods is based on prompting LLM to realize long-form story generation and the prompt is shown in Table~\ref{tab:exp_pronmpt}
\begin{figure*}[htbp]
\centering
\begin{tcolorbox}
Your task is to write a long-form story based on the input setting, character introduction and outline.\\
    The story should be consistent with the setting, character introduction and outline.\\
    The content of the story should be detailed and vivid, including detailed plot development, psychological description, environmental description, etc.\\
    \#\# Input\\
    the setting:{setting}\\
    the character introduction:{character}\\
    the outline:{outline}\\
    
    \#\# Output Format \\
    Nothing but only the generated content, which is composed by a series of sentences, should be included in the output.\\
    Strictly follow the format below:\\
    - Story: \\
    Your generated content:
\end{tcolorbox}
\caption{The prompt for generate story by LLMs.}
\label{tab:exp_pronmpt}
\end{figure*}

%
%
%
%
\section{Human evaluation details}
\label{sup:human evaluate}

\begin{figure*}[htbp]
\centering
\includegraphics[width=0.9\linewidth]{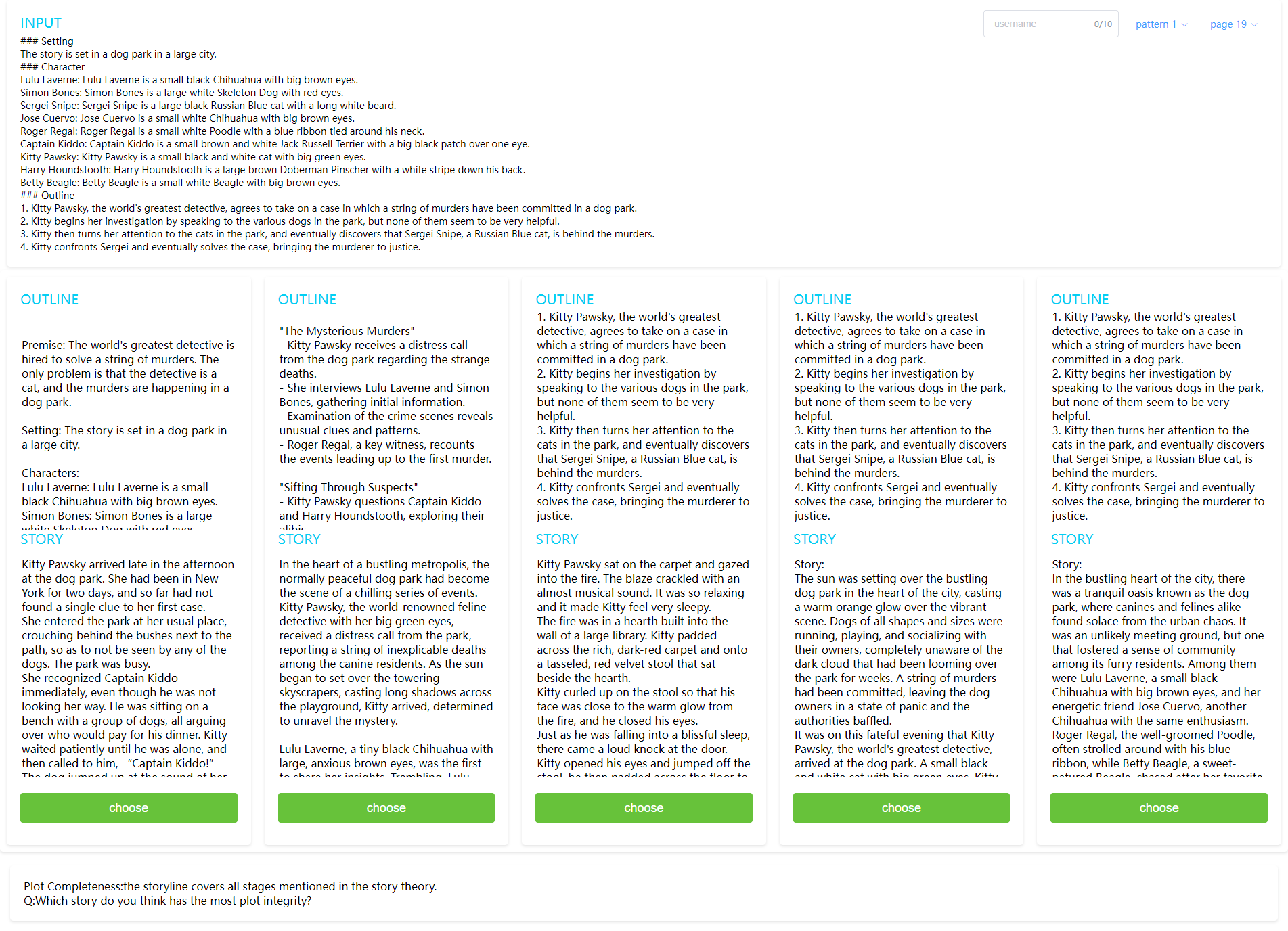}
\caption{The interface for human interface in comparison experiment.}
\label{fig:eval_path1}
\end{figure*}

\begin{figure*}[htbp]
\centering
\includegraphics[width=0.9\linewidth]{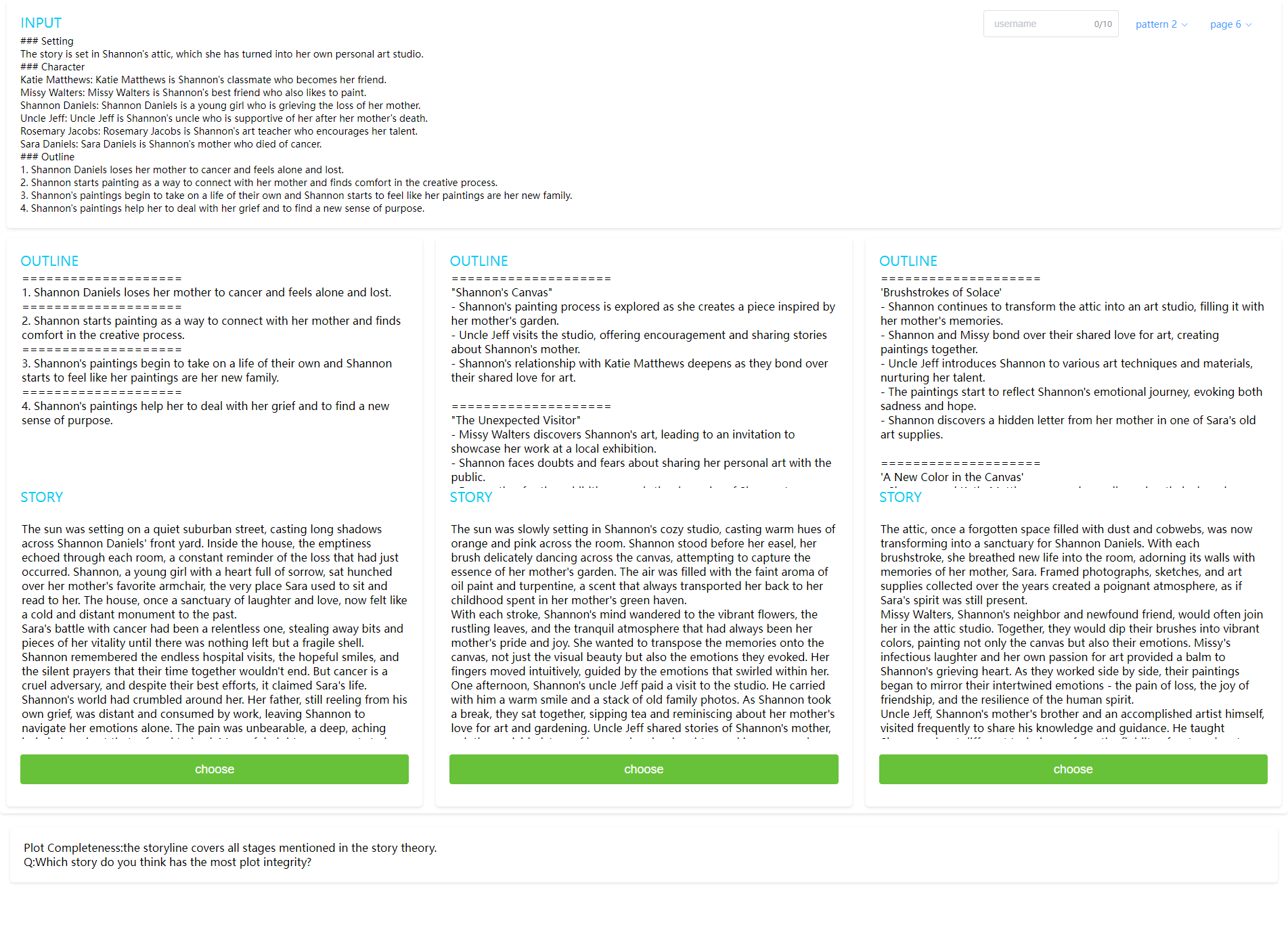}
\caption{The interface for human interface in ablation experiment.}
 \label{fig:eval_path2}
\end{figure*}

\begin{figure*}[htbp]
\centering
\includegraphics[width=0.9\linewidth]{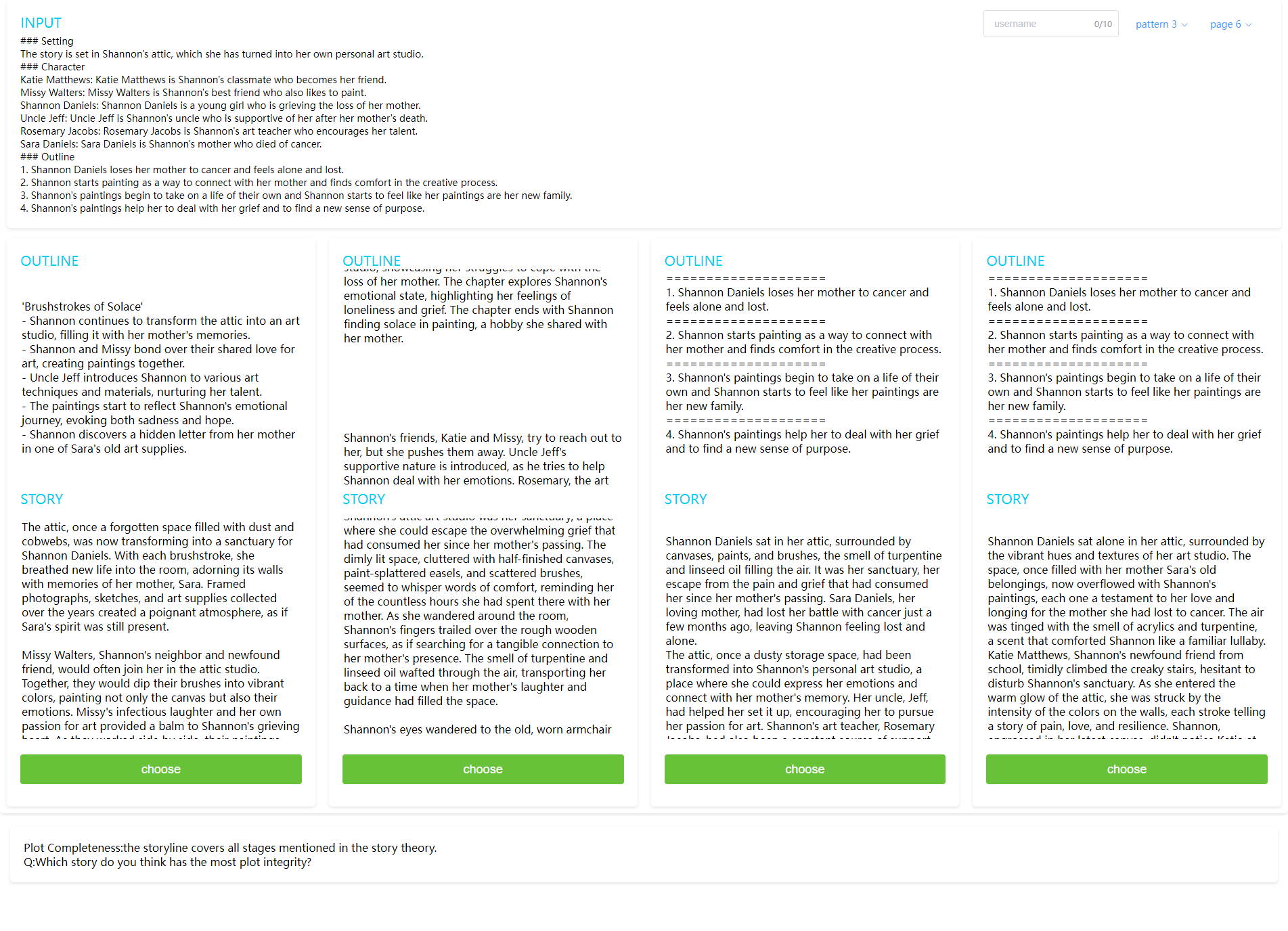} 
\caption{The interface for human interface in scalability experiment.}
\label{fig:eval_path3}
\end{figure*}

We describe the experimental details for human evaluation in the experiments mentioned in \ref{experiment}. For every evaluation, we prepare 20 groups of stories generated by different methods or experiment settings and ask human evaluators to rank stories in a group from best fit to least fit according to the given human metrics.
Specifically, we recruited three well-educated graduate students as evaluators and asked them to perform a blind ranking process. We designed a platform for the ranking process, where the interface displays input, generates detailed outlines, and generates stories from different methods or experiment settings. Fig.~\ref {fig:eval_path1}, Fig.~\ref{fig:eval_path2} and Fig.~\ref{fig:eval_path3} show the interface used for human evaluation in different experiments. Evaluators are required to read input, detailed outlines, and stories choose to the best stories in the current interface that fits the human metrics. Every chosen story disappears and its corresponding method is recorded. For every human metric, evaluators continue to choose the story best fits the metric in the current interface until there is no story shown in the interface.

\section{Grouping Rules}
\label{sup:grules}
Fig.~\ref{fig:grules} shows the grouping rules we applied to classify quadruples based on the knowledge expression features.
\begin{figure*}[h]
\centering
\includegraphics[width=0.90\linewidth]{images/grules.pdf}
    \caption{
    The rules for grouping quadruples.
    }  
    \label{fig:grules}
\vspace{-6pt}
\end{figure*}

\section{The example of output from MEM }
\label{sup:mem example}
\subsection{The example of extracted quadruples}
\label{sup:quadruples example}

Here is an example of quadruples which are constructed by story content and the result is shown in Fig.~\ref{fig:4ple_example}. Specifically, triples are
LLM is restricted to building triples from text, including sub,ac, and ob, and then the program adds the current chapter number to each triplet to form a quadruple. 

\begin{figure*}[h!]
\centering
\resizebox{0.85\linewidth}{!}{
\begin{tcolorbox}
\textbf{Text}\\
GabrielMartin, a young man in his early twenties, found himself nestled in the cozy confines of his newly acquired apartment. The transition to independence had been both exhilarating and daunting, but he had managed to keep his sanity amidst the flurry of change. One evening, as he sat by the window, watching the city lights twinkle in the distance, his thoughts drifted to Lina Sanchez, his girlfriend who was now miles away.Their long-distance relationship, once filled with excitement and promise, began to show the strain of their new living situations. The once effortless connection seemed to require more effort, their laughter now tinged with a hint of sadness. The silence between them on phone calls grew heavier, and it was during one such call that the tension finally broke."Lina, I can't ignore the fact that things feel different lately," Gabriel said, his voice filled with concern. "We're both adjusting to new lives, and it's hard not to feel the distance."Lina's voice, though soft, conveyed her own struggles. "I know, Gabriel. I miss you too, and sometimes it's overwhelming. But we love each other, and we'll find a way through this, won't we?"Gabriel's mind wandered back to his parents' relationship, their love having weathered storms far greater than the one they now faced. He remembered how they had maintained their bond even when separated by duty and distance. Angelica, his wise older sister, had once shared her insights on long-distance love, emphasizing the importance of communication and trust.Recalling Angelica's advice, Gabriel took a deep breath and spoke from the heart. "Lina, let's be honest with each other about our fears and insecurities. We need to make an effort to stay connected, even if it means having these tough conversations. Small gestures, like sharing our daily routines, can mean the world."Lina agreed, her voice gaining strength. "You're right, Gabriel. I'll try to be more open. And we'll plan visits, surprise each other, and make the distance worth it."Gabriel felt a weight lifted off his shoulders as they mapped out a plan to nurture their relationship. Their conversation, though difficult, brought them closer than ever. As they hung up, Gabriel looked at the glowing screen of his phone, a symbol of their determination to overcome the test of distance. In this new chapter of their lives, they would find a way to keep their love alive, no matter the obstacles they faced.\\
\\
\textbf{Quadruples}\\
<GabrielMartin, finds himself, nestled in newly acquired apartment, 2>\\
<transition, is, exhilarating and daunting, 2>\\
<Gabriel, manages to keep, sanity, 2>\\
<Gabriel, sits by the window, watching city lights, 2>\\
<Gabriel's thoughts, drift to, Lina Sanchez, 2>\\
<relationship, fills with, excitement and promise, 2>\\
<connection, seems to require, more effort, 2>\\
<laughter, tinged with, sadness, 2>\\
<silence, grows heavier, on phone calls, 2>\\
<Gabriel, speaks, with concern, 2>\\
<Gabriel, says, things feel different, 2>\\
<Gabriel, acknowledges, adjusting to new lives, 2>\\
<distance, is felt, by Gabriel, 2>\\
<Lina's voice, conveys, her struggles, 2>\\
<Lina, admits, missing Gabriel, 2>\\
<Gabriel, thinks of, parents' relationship, 2>\\
<Angelica, shares, insights on long-distance love, 2>\\
<Gabriel, remembers, importance of communication and trust, 2>\\
<Gabriel, takes a deep breath, speaks from the heart, 2>\\
<Gabriel, suggests, honesty about fears and insecurities, 2>\\
<Gabriel, proposes, staying connected, 2>\\
<Lina, agrees, to be more open, 2>\\
<Lina, plans, visits and surprises, 2>\\
<Gabriel, feels weight lifted off, shoulders, 2>\\
<conversation, brings, them closer, 2>\\
<Gabriel, looks at, glowing phone screen, 2>\\
<Gabriel and Lina, determined, to overcome distance, 2>\\
<they, plan, to keep love alive, 2>\\
<obstacles, they, will face, 2>\\
\end{tcolorbox}}
\caption{The example of quadruples.}
\label{fig:4ple_example}
\end{figure*}

\subsection{The example of provided relevant information}
\label{sup:relevant information example}
Here are some examples of the query and its relevant information provided by MEM. The examples are shown in Fig.~\ref{tab:rough_query} and Fig.~\ref{tab:detailed_query}. These results are relevant to the query since they contain the same entity and are semantically similar. In Fig.~\ref{tab:rough_query}, we show the query result about a rough outline. The information provided by \textbf{MEM} mentioned the same protagonists with the query, like "Brad" and "Shannon".Besides, the information provided by \textbf{MEM} gives more details about the query. For example, for"Brad's secretive behavior"  in the rough outline, there are details in the provided information "late nights at work and hurried phone calls" and there are more summarized details about the changes in the time dimension. In Fig.~\ref{tab:detailed_query}, we show the query result about a detailed outline. Consistent with the information for the rough outline, the information for the detailed outline provides more relevant details about the detailed outline, For example, for "Sergei Snipe's connection to the murders" in the detailed outline, there are details about connection "Sergei Snipe was observed close to crime scenes".

\begin{figure*}[htbp]
\centering
\begin{tcolorbox}
\textbf{Query:}\\
Brad's infidelity is gradually revealed through his secretive behavior and distant attitude. Shannon becomes increasingly suspicious and the tension in their household escalates. Jodi tries to support Shannon while Brad becomes more conflicted about his actions. Karen's presence in Brad's life becomes more prominent, creating conflict with Shannon.\\
\\
\textbf{Output:}\\
Brad had an affair with Karen Spooner.\\
Brad consumed memories related to his affair with Karen Spooner.\\
Brad made a terrible mistake involving Karen Spooner.\\
Brad's behavior changes, indicated by late nights at work and hurried phone calls.\\
The affair was caused by Brad.\\
Brad's guilt and pain are conveyed through silent tears.
Shannon is descending into darkness.\\
Shannon is considering extreme actions.\\
Shannon extends a communication or gesture towards Jodi.\\
Brad had an affair with Karen Spooner from chapter 1 to chapter 4.\\
Brad was consumed with memories of his affair with Karen Spooner from chapter 1 to chapter 4.\\
Brad made a terrible mistake with Karen Spooner from chapter 1 through chapter 4.\\
The affair was caused by Brad from chapter 1 to chapter 5.\\
Shannon reaches out to Jodi from chapter 3 to chapter 5.\\
\\
\end{tcolorbox}
\caption{The example of rough query of MEM.}
\label{tab:rough_query}
\end{figure*}

\begin{figure*}[htbp]
\centering
\begin{tcolorbox}
\textbf{Query:}\\
"Uncovering the Truth"\\
- Kitty Pawsky confronts Sergei Snipe and uncovers his connection to the murders.\\
- A chase ensues through the dog park and the city streets.
- A final confrontation takes place, revealing the true culprit and their motive.\\
- Kitty Pawsky explains the case to the relieved dog park inhabitants, restoring peace.\\
\\
\textbf{Output:}\\
Kitty Pawsky moved closer to revealing the truth.\\
Kitty Pawsky sets out to solve a case.\\
Kitty Pawsky solves a case.\\
Kitty Pawsky confronts Sergei Snipe.\\
Kitty Pawsky discovers Sergei Snipe.\\
Kitty Pawsky's whiskers are twitching.\\
Sergei Snipe was observed close to crime scenes.\\
The murders have left the animal community in a state of fear and unease.\\
The murders were committed in a dog park.\\
The dog park is situated within a large city.\\
Dogs seem to be not helpful.\\
The case involves a string of murders.\\
Kitty Pawsky is determined to protect the city.\\
Kitty Pawsky strategized in the quiet of the night.\\
Kitty Pawsky decided to keep a closer eye on Sergei Snipe.\\
Kitty Pawsky confronts Sergei Snipe from chapter 1 to chapter 2.\\
Kitty Pawsky discovers Sergei Snipe from chapter 1 to chapter 2.\\
Sergei Snipe was spotted near crime scenes from chapter 1 to chapter 2.\\
 Murders were committed in the dog park from chapter 1 to chapter 2.\\
The case involves a string of murders from chapter 1 to chapter 2.\\
Kitty Pawsky decided to keep a closer eye on Sergei Snipe from chapter 1 to chapter 2.\\
\end{tcolorbox}
\caption{The example of detailed query of MEM.}
\label{tab:detailed_query}
\end{figure*}

\section{The example of generated hierarchical outline in DHO}
\label{sup:dho example}
Here we post an example of a hierarchical outline including the rough outline and the detailed outline and made some analysis to validate the effectiveness of \textbf{DHO} module. The rough outline is shown in Fig.~\ref{fig:rough example} and the detailed outline Table~\ref{tab: detailed example}. 
\begin{figure*}[htbp]
\centering
\begin{tcolorbox}
\textbf{Stage: Exposition}\\
Gabriel Martin, a young man in his early twenties, moves out of his parents' house in a small town in the United States. Adjusting to life away from his mother, Angelica, father, Julio, and the familiar surroundings, he settles into his new apartment. Charles, his neighbor, offers a friendly welcome, and Gabriel's friends Richie and Jeremy lend a hand with the move. Meanwhile, Gabriel's relationship with Lina Sanchez undergoes the test of his newfound independence.\\
\\
\textbf{Stage: Rising Action}\\
Gabriel discovers the challenges of living on his own, such as noisy neighbors, unexpected maintenance issues, and budgeting. He faces conflicts with his landlord and has to learn how to balance his social life, work, and personal responsibilities. His friendships with Richie and Jeremy become strained due to their different lifestyles, while Lina starts to feel neglected by Gabriel's busy schedule.\\
\\
\textbf{Stage: Climax}\\
Gabriel's apartment catches fire due to an electrical malfunction, putting his life and belongings at risk. He narrowly escapes, but the incident escalates his conflicts with the landlord, who denies responsibility. This event becomes a turning point, forcing Gabriel to confront his inability to manage his life effectively and to seek help from his family and friends. Lina confronts Gabriel about their relationship, threatening to break up if he doesn't make time for her.\\
\\
\textbf{Stage: Falling Action}\\
Gabriel leans on his parents for support and learns from their experience, while also reconciling with Richie and Jeremy. Together, they rebuild his apartment and help him establish a more balanced routine. Gabriel apologizes to Lina and commits to making their relationship a priority. The landlord, under pressure from the community, agrees to compensate Gabriel for his losses.\\
\\
\textbf{Stage: Denouement or Resolution}\\
Gabriel's life stabilizes as he learns valuable lessons about responsibility, friendship, and love. He started a small support group for first-time apartment dwellers, helping others navigate the challenges he faced. His relationships with his family, friends, and Lina become stronger, and he finds a sense of belonging in his new community. The story concludes with Gabriel's growth and the resolution of all conflicts.\\
\end{tcolorbox}
\caption{The example of rough outline in our output.}
\label{fig:rough example}
\end{figure*}

\begin{small}
\onecolumn
\begin{longtable}{p{0.9\textwidth}}
\toprule
\textbf{Example Detailed Outline}\\
\midrule
\textbf{Brushstrokes of Solace}\\
Shannon continues to transform the attic into an art studio, filling it with her mother's memories.\\
Shannon and Missy bond over their shared love for art, creating paintings together.\\
Uncle Jeff introduces Shannon to various art techniques and materials, nurturing her talent.\\
The paintings start to reflect Shannon's emotional journey, evoking both sadness and hope.\\
Shannon discovers a hidden letter from her mother in one of Sara's old art supplies.\\
\\
\textbf{A New Color in the Canvas}\\
Shannon and Katie Matthews grow closer, discussing their shared interests and dreams.\\
Katie, inspired by Shannon's art, starts painting herself, discovering a hidden talent.\\
The paintings in the attic exhibit an inexplicable energy, hinting at a connection to Sara's spirit.\\
Shannon, Missy, and Katie have a joint art exhibition at school, showcasing their emotional growth.\\
Shannon receives a positive response from her peers, helping her find a sense of belonging.\\
\\
\textbf{Whispers from the Canvas}\\
Shannon starts to sense her mother's presence through the paintings, sensing guidance and comfort.\\
Uncle Jeff, noticing the changes in the artwork, shares his belief in the spiritual connection.\\
Shannon, Missy, and Katie have a late-night painting session, where they all feel their loved ones' presence.\\
Shannon finds closure in her grief, understanding that her mother's love is with her always.\\
The chapter ends with Shannon painting a vibrant, uplifting piece, symbolizing her newfound hope.\\
\\
\textbf{The Mysterious Brushstrokes}\\
Shannon, Katie, and Missy start investigating the phenomenon, visiting libraries and seeking advice from art experts.\\
Shannon has a revealing conversation with Rosemary Jacobs about the potential supernatural aspects of her art.\\
The trio experiments with Shannon's painting techniques, trying to recreate the living effect.\\
A late-night encounter with one of Shannon's paintings leaves them more confused and intrigued.\\
\\
\textbf{A Glimpse Beyond the Canvas}\\
Shannon has a vivid dream where her mother communicates through the paintings.\\
The girls attend an art seminar where they discuss the concept of energy and its influence on art.\\
Uncle Jeff shares stories of his own unexplained experiences, providing a broader perspective.\\
Shannon starts to suspect that her mother's spirit may be channeling through her art.\\
A heartfelt conversation with Katie and Missy solidifies their bond and their dedication to understanding the mystery.\\
\\
\textbf{The Living Gallery}\\
Shannon, Katie, and Missy decide to showcase Shannon's enchanted paintings in a small gallery exhibit.\\
Visitors react to the paintings in unexpected ways, experiencing emotions and memories triggered by the art.
Rosemary Jacobs attends the exhibit, offering insights and support.\\
Shannon receives a mysterious note from someone claiming to know the truth behind the phenomenon.\\
The chapter ends with Shannon feeling a mix of excitement and apprehension, eager to uncover the secrets behind her artwork.\\
\\
\textbf{Unraveling the Message}\\
Shannon studies the hidden message within her painting, seeking to decipher its meaning.\\
With Katie and Missy's help, Shannon investigates her mother's past for clues.
Shannon visits Rosemary Jacobs for guidance, learning more about Sara's history and her own supernatural gift.
A confrontation with skepticism from those around her, testing Shannon's resolve.\\
\\
\textbf{The Choice Looms}\\
Shannon grapples with the implications of embracing her gift fully or suppressing it for a "normal" life.\\
The strain on her relationships with friends and family is highlighted, especially with her father's disapproval.\\
Shannon has a heart-to-heart conversation with Katie about the importance of truth and acceptance.\\
A dream sequence reveals more about Sara's intentions and the importance of Shannon's role.\\
\\
\textbf{Embracing the Unknown}\\
Shannon decides to embrace her supernatural ability, understanding the impact it can have on others.\\
She begins to integrate her gift into her art, creating new works that offer healing and understanding.\\
A public demonstration of her gift brings both support and skepticism, solidifying Shannon's resolve.\\
The chapter ends with Shannon feeling a newfound sense of purpose and connection to her mother's memory.\\
\\
\textbf{The Healing Canvas}\\
Shannon continues to develop her unique art therapy, using her ability to connect with mourners through her paintings.\\
She holds a workshop with a group of grieving individuals, including Katie and other classmates, demonstrating the healing power of her art.\\
Shannon's mother's essence is felt by all, fostering a sense of unity and understanding among the participants.\\
Uncle Jeff and friends support Shannon in her newfound mission, helping her organize more events.\\
\\
\textbf{A Community's Response}\\
The local community becomes aware of Shannon's talent, and she receives invitations to share her gift at different gatherings.\\
Shannon learns to navigate both the admiration and skepticism, strengthening her conviction in her purpose.\\
She faces a challenging session with a skeptic, who eventually experiences a breakthrough, validating Shannon's approach.\\
The town begins to heal collectively, with Shannon's art playing a central role.\\
\\
\textbf{The Letter's Revelation}\\
Shannon discovers a hidden letter from her mother in one of Sara's old paintings, providing a personal message of love and encouragement.\\
The letter deepens Shannon's understanding of her mother's love and her own destiny.\\
Shannon reads the letter publicly at an event, connecting with the audience on a deeper emotional level.\\
The story concludes with Shannon finding closure and embracing her role as a healer, solidifying her bond with her mother's memory.\\
\\
\textbf{Preparation for Hope}\\
Shannon finalizes her artwork for the exhibition, incorporating her emotional journey and her mother's spirit.\\
Shannon reflects on her growth and the impact of Sara's love on her art.\\
Katie and Missy offer support and encouragement, strengthening their friendship.\\
Shannon's father starts to show interest in her art, hinting at a shift in their relationship.\\
\\
\textbf{Exhibition of Resilience}\\
The art exhibition opens, with Shannon's paintings attracting attention and evoking emotions in visitors.\\
Shannon shares her story, inspiring others to find hope in their own struggles.\\
Shannon has a heartfelt conversation with her father, who acknowledges her talent and experiences.\\
Shannon, Katie, and Missy share a bonding moment, solidifying their friendship bond.\\
\\
\textbf{A New Beginning}\\
Shannon receives positive feedback on her work, boosting her confidence as an artist.\\
She reads the hidden letter from her mother, which brings a sense of closure and understanding.\\
Shannon decides on her artistic path, embracing her destiny to share hope through her art.\\
The chapter concludes with Shannon, Katie, and Missy standing together, symbolizing unity and hope for the future.\\

\textbf{New Beginnings}\\
Gabriel settles into his new apartment, experiencing the initial excitement and anxiety of independence.\\
He interacts with his neighbor, Charles, who extends a warm welcome and offers advice about apartment living.\\
Richie and Jeremy visit Gabriel to help with unpacking, reminiscing about past adventures and shared memories.\\

\textbf{The Test of Distance}\\
Gabriel's long-distance relationship with Lina Sanchez starts to show strain due to their new living situations.\\
They have a heartfelt conversation over the phone, discussing their concerns and how to maintain their connection.\\
Gabriel reflects on his parents' relationship and Angelica's advice on handling long-distance love.\\

\textbf{Navigating Challenges}\\
Gabriel faces unexpected challenges of apartment living, such as noisy neighbors and learning to manage household tasks.\\
He turns to Charles for guidance, who shares his own experiences and offers practical solutions.\\
Gabriel's independence grows as he troubleshoots these problems, strengthening his resolve and self-reliance.\\

\textbf{Budgeting Blues}\\
Gabriel realizes the financial realities of independent living, encountering unexpected expenses and budget constraints.\\
He attempts to economize, cutting back on non-essentials and searching for ways to increase his income.\\
The strain of budgeting causes tension between Gabriel and his friends, as they indulge in activities he can no longer afford.\\

\textbf{Friendship Friction}\\
Richie and Jeremy's carefree lifestyle contrasts with Gabriel's newfound responsibilities, leading to misunderstandings.\\
Gabriel feels resentment over their lack of understanding, while Richie and Jeremy struggle to empathize with his situation.\\
A confrontation between Gabriel and his friends highlights the growing rift in their relationships.\\

\textbf{Reconnecting with Lina}\\
Gabriel acknowledges Lina's feelings of neglect and makes an effort to prioritize their relationship.\\
They have a heart-to-heart conversation about the challenges they both face, seeking mutual understanding.\\
A plan is devised to balance Gabriel's responsibilities with quality time for the couple, setting the stage for mending their bond.\\

\textbf{In the Wake of Disaster}\\
Gabriel grapples with the aftermath of the apartment fire, sorting through the remnants of his belongings.\\
He meets with the landlord to discuss the incident, leading to a heated argument over responsibility.\\
Gabriel's frustration deepens as he realizes the legal implications and potential financial burden.\\

\textbf{Seeking Support}\\
Gabriel turns to his family for help, sharing his struggles and the recent events with them.\\
They offer emotional support and advice on how to handle the situation with the landlord.\\
Gabriel starts exploring temporary housing options, with the assistance of his family and friends.\\

\textbf{Lina's ultimatum}\\
Lina visits Gabriel amidst the chaos, expressing her concern for both their relationship and his well-being.\\
The fire serves as a backdrop for their conversation, heightening the urgency for change.\\
Gabriel commits to addressing not only the fire-related issues but also to making their relationship a priority, as threatened by Lina's breakup warning.\\

\textbf{Rebuilding and Reconciliation}\\
Gabriel, Richie, and Jeremy continue the repairs on Gabriel's apartment, turning it into a welcoming home.\\
Gabriel and Richie mend their friendship, discussing the past misunderstandings and finding common ground.\\
Jeremy offers emotional support, sharing his own experiences with overcoming challenges.\\

\textbf{A New Routine and Priorities}\\
Gabriel establishes a weekly cleaning and self-care schedule to maintain balance in his life.\\
He meets with Lina to discuss their relationship, openly addressing his fears and insecurities.\\
Lina and Gabriel set goals for strengthening their connection, overcoming the distance created by the recent events.\\

\textbf{New Routine and Priorities}\\

Gabriel establishes a weekly cleaning and self-care schedule to maintain balance in his life.\\

He meets with Lina to discuss their relationship, openly addressing his fears and insecurities.\\

Lina and Gabriel set goals for strengthening their connection, overcoming the distance created by the recent events.\\

\textbf{Compensation and Community Support}\\

The landlord, influenced by the community's pressure, meets with Gabriel to discuss compensation for his losses.\\

Gabriel negotiates a fair deal, showing maturity and resilience in the face of adversity.\\

The chapter ends on a positive note as Gabriel's relationships with both Lina and his friends deepen, signaling a new beginning.\\

\textbf{A New Beginning}\\
Gabriel and Lina solidify their commitment to each other, focusing on their deepening love.\\
Gabriel starts the support group, sharing his experiences and lessons learned with others.\\
The group dynamics and friendships formed within the support group are described.\\
Gabriel's relationship with his family improves, as they see his growth and responsibility.\\

\textbf{Growth and Resolution}\\
Gabriel confronts and resolves any lingering conflicts with his friends, showcasing his maturity.\\
Lina's trembling subsides as she finds comfort in Gabriel's support and open communication.\\
The couple faces and conquers a specific challenge together, demonstrating their newfound strength as a team.\\
The community's acceptance of Gabriel and Lina is highlighted, fostering a sense of belonging.\\

\textbf{Celebrating Unity}\\
A community event is organized, where Gabriel's support group is recognized and celebrated.\\
Gabriel's landlord surprises him with a gesture of appreciation, reflecting the positive change in their relationship.\\
Gabriel and Lina share a special moment, symbolizing their growth and love.\\
The volume concludes with a look at the future, as Gabriel and Lina plan their next steps together, ready for whatever challenges lie ahead.\\

\bottomrule
\caption{The example of detail outline. }
\label{tab: detailed example}
\end{longtable}
\end{small}

\section{The example of generated story}
\label{sup:story example}
There is an example of a generated story which is shown in Table~\ref{tab: story example1}. These results were generated based on Qwem1.5-72B-chat.We make a case study of the story example shown in Table~\ref{tab: story example1}.
Its hierarchical outline of the story is shown in Fig.~\ref{fig:rough example} and Table~\ref{tab: detailed example}.\par
While there are still some issues in this story, it shows coherence in plot and expression. The rough outline is complete and it covers all stages mentioned in the novel writing theory. 
Besides, the development of the story and the character description are consistent with the input, indicating high relevance to the input story premise.
The story mentions Gabriel's girlfriend is in a different place which is the creation during the generation stage and the subsequent follows this statement, indicating the story development fluency due to the adaption to the uncertainty of the wertting stage.
In the example story, the plots from Gabriel encounter challenges to solve them have
expanded in detail, explaining the details of reasonable solutions and the status changes of Characters. As the story proceeds, many difficulties are overcome by Gabriel, improving the readability and attraction to readers. 
His neighbor appeared from the beginning to the end with a consistent expression like"wise oldest man", "the wisdom of his neighbor", "older gentleman with a friendly smile" and "Charles listened intently, nodding empathetically". This expression is consistent, indicating the contextual consistency ensured through the relevant content provided in the writing stage by MEM. 
Therefore, the example story can demonstrate all the features that DOME attempts to improve and indicate the effectiveness of DOME.

\begin{small}
\onecolumn
\begin{longtable}{p{0.9\textwidth}}
\toprule
\textbf{Example Story}\\
\midrule
Gabriel Martin, a sprightly young man with a touch of nervousness in his eyes, stepped into his new apartment for the very first time. The space was modest, but it represented a significant milestone in his life - the first step towards independence. The walls, still bare, echoed the silence of the uncharted territory he had just entered. A mix of excitement and anxiety pulsed through him, his heart thumping with anticipation.
\\
\\
As he stood in the middle of the room, taking in the blank canvas before him, a knock on the door startled him out of his thoughts. Peering through the peephole, he saw an older gentleman with a friendly smile - his neighbor, Charles. Charles extended a warm welcome, his weathered hand engulfing Gabriel's in a reassuring grip. "Welcome to the building, Gabriel. If you ever need anything, just give me a holler." His kind demeanor instantly put Gabriel at ease.
\\
\\
Over a cup of steaming coffee, Charles shared his wisdom about apartment living - from the quirks of the building's heating system to the best local grocery store. Gabriel listened intently, grateful for the guidance. Charles's stories of his own experiences navigating independence made Gabriel feel less alone in his new adventure.
\\
\\
A few days later, the doorbell rang again, this time bringing a burst of energy and laughter. Richie and Jeremy, Gabriel's childhood friends, had arrived to lend a hand with unpacking. The room was soon filled with a jumble of boxes and a flurry of nostalgia. As they worked, they reminisced about their shared memories - from the time they built a fort in Gabriel's backyard to their wild teenage escapades.
\\
\\
Richie, the charismatic charmer, regaled them with tales of his recent travels, while Jeremy, the introverted thinker, shared his latest philosophical musings. Their banter filled the apartment with a familiar warmth, easing Gabriel's anxieties about his new life. Together, they laughed, reminisced, and forged ahead, turning the empty apartment into a place that felt like home.
\\
\\
Gabriel's new beginnings were marked by challenges, but with the support of his friends and the wisdom of his neighbor, he was learning to navigate the complexities of apartment living. Each day was a new adventure, and he was determined to embrace every moment of it, keeping his sanity intact and his heart open to the experiences that lay ahead.
\\
\\
Gabriel Martin, a young man in his early twenties, found himself nestled in the cozy confines of his newly acquired apartment. The transition to independence had been both exhilarating and daunting, but he had managed to keep his sanity amidst the flurry of change. One evening, as he sat by the window, watching the city lights twinkle in the distance, his thoughts drifted to Lina Sanchez, his girlfriend who was now miles away.
\\
\\
Their long-distance relationship, once filled with excitement and promise, began to show the strain of their new living situations. The once effortless connection seemed to require more effort, their laughter now tinged with a hint of sadness. The silence between them on phone calls grew heavier, and it was during one such call that the tension finally broke.
\\
\\
"Lina, I can't ignore the fact that things feel different lately," Gabriel said, his voice filled with concern. "We're both adjusting to new lives, and it's hard not to feel the distance."
\\
\\
Lina's voice, though soft, conveyed her own struggles. "I know, Gabriel. I miss you too, and sometimes it's overwhelming. But we love each other, and we'll find a way through this, won't we?"
\\
\\
Gabriel's mind wandered back to his parents' relationship, their love having weathered storms far greater than the one they now faced. He remembered how they had maintained their bond even when separated by duty and distance. Angelica, his wise older sister, had once shared her insights on long-distance love, emphasizing the importance of communication and trust.
\\
\\
Recalling Angelica's advice, Gabriel took a deep breath and spoke from the heart. "Lina, let's be honest with each other about our fears and insecurities. We need to make an effort to stay connected, even if it means having these tough conversations. Small gestures, like sharing our daily routines, can mean the world."
\\
\\
Lina agreed, her voice gaining strength. "You're right, Gabriel. I'll try to be more open. And we'll plan visits, surprise each other, and make the distance worth it."
\\
\\
Gabriel felt a weight lifted off his shoulders as they mapped out a plan to nurture their relationship. Their conversation, though difficult, brought them closer than ever. As they hung up, Gabriel looked at the glowing screen of his phone, a symbol of their determination to overcome the test of distance. In this new chapter of their lives, they would find a way to keep their love alive, no matter the obstacles they faced.
\\
\\
Gabriel Martin had always imagined apartment living as a glamorous, carefree existence, but reality quickly proved him wrong. The walls of his cozy one-bedroom apartment seemed paper-thin, amplifying the sounds of his neighbors' late-night parties and early-morning commotion. The once-appealing idea of independence was now tinged with frustration, as he found himself struggling to adapt to his new environment.
\\
\\
One afternoon, Gabriel decided to seek advice from his wise friend, Charles, an elderly gentleman who lived on the same floor. Charles had lived in the building for decades and was known for his wealth of knowledge and life experiences. Over cups of steaming tea in Charles's well-worn living room, Gabriel shared his woes about the noisy neighbors and his own struggles with managing household chores.
\\
\\
Charles listened intently, nodding empathetically. "When I first moved in," he began, "I, too, faced similar challenges. But with time, I learned a few tricks to maintain my peace of mind." He offered Gabriel practical tips, like using noise-cancelling headphones during the day and investing in thick curtains to block out unwanted sounds at night. He also shared his methods for staying organized and tackling household tasks efficiently.
\\
\\
Armed with Charles's wisdom, Gabriel returned to his apartment with a newfound determination. He implemented Charles's suggestions, filling his apartment with potted plants to absorb sound and establishing a weekly cleaning schedule. Each time he successfully navigated a new challenge, Gabriel felt a surge of pride and independence. His resolve strengthened, and he began to see these obstacles not as setbacks, but as opportunities for growth.
\\
\\
As the days turned into weeks, Gabriel's self-reliance blossomed. He formed connections with his neighbors, even the noisy ones, finding common ground and understanding. Their late-night parties became chances to bond over shared music tastes, turning a once-irritating situation into a source of community.
\\
\\
Gabriel's journey was far from over, but he was learning that navigating life's challenges was as much about adapting as it was about resilience. With each obstacle he overcame, he became more grounded in his new life, grateful for the support of friends and the wisdom of Charles that guided him along the way.
\\
\\
Gabriel sat at the small, worn kitchen table in his new apartment, staring at a pile of bills and receipts spread out before him. The once exciting thrill of independence had given way to a sobering realization – life on his own came with a price tag he hadn't fully anticipated. Rent, utilities, groceries, and the seemingly endless list of little expenses had added up quickly. The budget he'd naively drawn up before moving out now seemed hopelessly inadequate.
\\
\\
He picked up a flyer for a local gym, a membership he'd indulged in shortly after moving out. It seemed like a luxury now, a frivolous expense that needed to be cut. Sighing, he made a mental note to cancel it later. Next, he turned to his food expenses. Eating out, once a weekly treat, was now a thing of the past. He began scouring the internet for recipes that were both cheap and nutritious.
\\
\\
As days turned into weeks, Gabriel found himself working longer hours at the bookstore, hoping to boost his income. He started selling some of his old possessions online, items that once held sentimental value but were now deemed unnecessary. The apartment felt emptier, but his wallet felt slightly fuller.
\\
\\
The strain of budgeting began to seep into his social life. His friends, Charles included, still expected the same Gabriel who could afford to join them for drinks after work or spontaneous weekend getaways. Every time they made plans, Gabriel felt a pang of guilt and sadness. He couldn't afford to be a part of their adventures anymore, and he hated the idea of letting them down.
\\
\\
Charles, sensing Gabriel's growing distance, confronted him one evening after work. "Man, what's going on with you lately? You've been distant, always finding excuses not to hang out," Charles said, concern etched on his face.
\\
\\
Gabriel hesitated, unsure of how to explain his newfound financial struggles without appearing weak. Then he remembered his own advice – honesty was key. He took a deep breath and shared his budgeting woes with Charles, hoping his friend would understand.
\\
\\
Charles nodded empathetically, "I get it, Gabe. Money's tight for all of us at some point. We'll find ways to make it work. Maybe we can have game nights at your place or go for hikes instead of hitting up bars."
\\
\\
Gabriel felt a wave of gratitude, but the shadow of Lina Sanchez still lingered in his thoughts. He wondered how their long-distance relationship would fare under the weight of his budgeting blues. Despite the challenges, he remained determined to find a way to make it work, both for himself and for Lina. After all, adjusting to this new life was just another obstacle to overcome, and he was not one to back down easily.
\\
\\
The air in Gabriel's small apartment was heavy with the tension of unspoken words, a stark contrast to the usual carefree atmosphere that accompanied Richie and Jeremy's visits. The two friends, always up for an adventure, seemed oblivious to the changes in Gabriel's life. They laughed and joked as they helped with the unpacking, their laughter echoing through the empty rooms, amplifying Gabriel's newfound responsibilities.
\\
\\
Gabriel had been working longer hours at the bookstore, sacrificing his leisure time to make ends meet. The decision to sell his old possessions weighed heavily on his heart, each item a reminder of a life he was leaving behind. He had confided in Richie and Jeremy about his budgeting struggles, hoping for understanding, but their lightheartedness only served to deepen the divide.
\\
\\
Feeling resentment simmering beneath the surface, Gabriel watched them, envious of their freedom. They seemed to have no worries, no responsibilities holding them back. The more he observed, the more he wished they could comprehend the gravity of his situation. But how could he make them understand when they were still living in a world of carefree whims?
\\
\\
One evening, as the three friends sat around the modest dinner table, the silence between them grew too loud to ignore. Gabriel's frustration erupted in a burst of pent-up emotion. "You have no idea what it's like," he said, his voice shaking. "I'm trying to keep my head above water here, but it feels like I'm drowning."
\\
\\
Richie and Jeremy exchanged a surprised look, their easy smiles faltering. They hadn't realized the depth of Gabriel's struggles or the extent to which their lifestyle was rubbing salt in his wounds. Attempting to bridge the gap, Jeremy spoke softly, "We didn't mean to hurt you, Gabe. We just... we didn't know."
\\
\\
Gabriel's eyes met Jeremy's, and for a moment, he saw the genuine concern in his friend's gaze. The weight of their misunderstanding began to lift, but the rift was still palpable. It was a confrontation that marked the beginning of a difficult journey for all three, one where they would have to learn to balance their friendship with the realities of growing up.
\\
\\
In the midst of this turmoil, Gabriel couldn't help but think of Charles, the kind-hearted man he had recently met at the bookstore. Charles, with his wisdom and empathy, seemed to understand the complexities of life in a way that Richie and Jeremy couldn't yet grasp. Gabriel wondered if his new acquaintance could provide the guidance he needed to navigate these turbulent waters of friendship and responsibility.
\\
\\
Gabriel sat across from Lina at their cozy kitchen table, the warm glow of the overhead light casting a soft halo around her. He could see the exhaustion etched on her face, the shadows beneath her eyes a testament to the strain their recent distance had caused. He knew he had been neglectful, his longer hours at the bookstore leaving little room for the intimacy and connection they once shared.
\\
\\
"I've been aware of how I've been neglecting our relationship, Lina," Gabriel began, his voice filled with sincerity. "I've been so focused on work that I've let our bond slip away. That was a mistake, and I'm sorry."
\\
\\
Lina looked up at him, her own voice laced with hurt. "It's not just the time, Gabriel. It's the feeling that I'm not a priority anymore. That our problems and fears don't matter as much as the books."
\\
\\
Gabriel took a deep breath, remembering the importance of honesty and trust he had often preached. "You're right. I've been so consumed by the bookstore, trying to make it successful, that I've forgotten the foundation of our relationship. We need to talk about our struggles and support each other."
\\
\\
They delved into a heart-to-heart conversation, discussing their fears and insecurities, the pressures they both faced in their individual lives. Lina shared her longing for the simple moments they used to cherish, while Gabriel opened up about the mounting responsibilities at work.
\\
\\
Together, they devised a plan. Gabriel would set clear boundaries for his work hours, dedicating specific days and evenings for just the two of them. They would plan regular date nights, surprise each other with small gestures, and make a conscious effort to communicate daily, no matter how busy they were.
\\
\\
As the night wore on, the tension between them began to dissipate, replaced by a renewed sense of understanding and commitment. Gabriel vowed to prioritize their relationship, not just in words but in action, and Lina's eyes shone with hope, knowing that their bond was on the path to mending.
\\
\\
With a new determination, they held hands, their fingers entwining in a promise of dedication and love. Gabriel felt a wave of relief wash over him, grateful for the chance to reconnect with Lina and rekindle the spark that had brought them together.
\\
\\
In the wake of the devastating apartment fire, Gabriel found himself sifting through the charred remains of his once-tidy life. The acrid smell of smoke still lingered in the air, a constant reminder of the disaster that had befallen him. As he picked up a blackened photo frame, the edges crumbled under his touch, revealing a barely recognizable snapshot of happier times. The weight on his heart grew heavier with each item he salvaged.
\\
\\
The landlord, a stern man with a thick mustache, met Gabriel the following day in what had become a desolate landscape of debris. They stood amidst the ruins, the words between them as cold as the ashes beneath their feet. Gabriel argued that the building's outdated wiring had been the cause, while the landlord stubbornly insisted it was Gabriel's negligence. The exchange quickly escalated into a heated argument, with accusations and defenses flying back and forth.
\\
\\
Gabriel's frustration mounted as the conversation delved into the legal complexities of the situation. He realized that he might be held responsible for damages, a prospect that left him reeling. The potential financial burden loomed over him like a dark cloud, casting doubt on his ability to recover from this disaster.
\\
\\
As the reality of his circumstances set in, Gabriel's guilt and sadness intertwined with a sense of determination. He had neglected some basic safety precautions, and now he was paying the price. The guilt ate at him, but he knew he had to move forward. Friends and neighbors, their own homes spared from the fire, offered words of comfort and help in the form of advice and assistance with selling his possessions.
\\
\\
Gabriel's apartment, once a sanctuary, now felt cavernous and empty. But with each item he sold, he found a small sense of relief, as the slightly fuller wallet brought a glimmer of hope amidst the strain of budgeting. He knew the road ahead would be challenging, but the outpouring of support from those around him gave him the strength to face the journey.
\\
\\
In the midst of the chaos, Gabriel found gratitude in the unexpected kindness of others. It was a bittersweet realization, one that taught him the value of community and the fragility of life. The fire had taken so much from him, but it had also illuminated the resilience of his spirit and the enduring bonds he shared with those around him.
\\
\\
Gabriel sat in the warm glow of his parents' living room, the familiar surroundings providing a small comfort amidst the turmoil in his life. The weight of recent events rested heavily on his heart, but he knew he couldn't bear it alone anymore. With a deep breath, he began to share the story of the fire, the accusations from the landlord, and the overwhelming sense of guilt that had been haunting him.
\\
\\
His parents listened intently, their faces etched with concern. His mother's hand found his, offering silent reassurance, while his father's eyes reflected a mixture of anger and sympathy. They didn't doubt Gabriel's account, and their belief in him was a balm to his soul.
\\
\\
"You know, Gabriel," his father said, his voice firm, "nobody can blame you for something that's out of your control. You've always been responsible, and this is just an unfortunate incident. Now, let's focus on dealing with the situation."
\\
\\
His mother chimed in, "We'll help you find a way through this. In the meantime, consider staying with us for a while. It'll give you time to sort things out without the added stress of finding a new place right away."
\\
\\
Gabriel felt a wave of gratitude wash over him, grateful for their unwavering support. Friends from the neighborhood had also reached out, offering their assistance. Their kindness made him realize he wasn't as alone as he had thought.
\\
\\
With newfound determination, Gabriel started researching temporary housing options, aided by his family and friends. They scoured online listings and local community boards, looking for affordable places that would accept short-term tenants. His best friend, Leo, even suggested they pool their resources to rent a small house together until Gabriel got back on his feet.
\\
\\
As the days passed, Gabriel found himself growing stronger, buoyed by the love and support around him. The guilt and resentment began to dissipate, replaced by a sense of resilience and hope. He knew that the road ahead would be challenging, but with his family and friends by his side, he was ready to face it head-on.
\\
\\
In the midst of the chaos, Lina stepped into the heart of the fire zone, her eyes searching for Gabriel amidst the swirling smoke and the roar of the flames. The once peaceful neighborhood was now an inferno, with firefighters battling to contain the blaze that threatened to consume everything in its path. Lina's heart pounded with fear for Gabriel and their relationship, which had been hanging by a thread.
\\
\\
Gabriel appeared, his face smudged with soot and his eyes filled with concern. He noticed Lina's trembling form and rushed to her side, enveloping her in a protective embrace. The flickering orange and red light from the burning buildings cast an eerie glow on their faces, amplifying the gravity of their conversation.
\\
\\
"I can't do this anymore, Gabriel," Lina said, her voice wavering with emotion. "The fire, the stress, the neglect—it's all taking a toll on us. I want us to be a priority, not just another item on your to-do list."
\\
\\
Gabriel's gaze softened as he looked into Lina's eyes. He knew she was right, that their bond had suffered in the face of his preoccupations. The fire, a symbol of the neglect and pitfall that had been festering in their lives, illuminated the urgency for change.
\\
\\
"I promise, Lina," he vowed, his voice sincere and resolute. "I'll not only address the fire safety issues in our community but also in our relationship. We'll make time for each other, and I'll be more open about my fears and insecurities. Together, we'll work through this."
\\
\\
Lina's voice, still tinged with hurt, softened at Gabriel's words. She remembered the longing she had expressed for simple moments, for the intimacy they once shared. It wasn't about the quantity of time, but the quality.
\\
\\
"I believe you, Gabriel," she said, her hand finding his. "But actions speak louder than words. Let's start by being honest and rebuilding what we've lost."
\\
\\
With a renewed sense of commitment, Gabriel and Lina stood together, their hands clasped tightly, as the firefighters battled the blaze. The fire raged on, but in their hearts, a spark of hope ignited, promising a future where love and safety would coexist.
\\
\\
Gabriel, Richie, and Jeremy worked diligently to restore the small apartment to its former glory, turning the once-damaged space into a warm and inviting home. The scent of fresh paint and sawdust filled the air as they moved from one task to another, laughter and conversation easing the tension that had built up between them.
\\
\\
Gabriel and Richie, brushes in hand, paused to discuss the misunderstandings that had strained their friendship. Gabriel explained the fear and insecurity that had kept him from being open about his struggles, while Richie admitted to his own obliviousness. They found common ground in their shared experiences of life's challenges, and a new understanding began to form between them.
\\
\\
Jeremy, sensing the moment, softly shared his own story of overcoming adversity. His words were a balm to Gabriel's soul, reminding him that they were all in this together. The three friends, now bonded by their vulnerabilities and the effort to rebuild, created an atmosphere of healing and support.
\\
\\
As the evening light filtered through the newly replaced windowpanes, the apartment took on a cozy glow. The once-bare walls were adorned with artwork, and the worn furniture had been restored to its former charm. Gabriel, looking around at the fruits of their labor, felt a profound sense of accomplishment and gratitude.
\\
\\
The weight of recent events still lingered, but with each passing moment, it felt lighter. The guilt and resentment began to fade, replaced by a resolve to face the future with hope and resilience. The emptiness of the apartment was now filled not just with physical objects, but with the warmth of friendship and a renewed sense of purpose.
\\
\\
Gabriel, Richie, and Jeremy sat together on the couch, a comfortable silence settling between them. There was a newfound peace in the air, one that spoke of reconciliation and the promise of a better tomorrow. As they exchanged warm smiles, Gabriel felt a wave of relief wash over him, knowing that he was no longer alone in his journey. The stress of budgeting and the weight of responsibility still existed, but now, it was tempered by the love and support of his friends.
\\
\\
With the sun beaming through his window, Gabriel started his day with a newfound determination. He sat at his kitchen table, penning out a weekly cleaning and self-care schedule, carefully allocating time for exercise, meditation, and quality moments with Lina. He understood that maintaining balance in his life was crucial, not just for himself but also for the wellbeing of their relationship.
\\
\\
Gabriel called Lina, inviting her over for a heart-to-heart conversation. As she arrived, her warm smile filled the room, but Gabriel could sense the underlying tension. They sat on the couch, their hands intertwined, as he began to share his insecurities. He spoke about the fear of losing her due to his recent neglect and the anxiety that sometimes overwhelmed him. Lina listened patiently, her eyes filled with empathy.
\\
\\
In return, Lina shared her own thoughts, acknowledging the strain recent events had put on them. She admitted to feeling distant from Gabriel but expressed her willingness to work through it together. They then set joint goals to strengthen their bond, such as having regular date nights, engaging in deep conversations, and supporting each other's personal growth.
\\
\\
The air in the room seemed to lighten as they talked, the weight of their concerns slowly lifting. Gabriel felt a renewed sense of hope, remembering the importance of communication and trust, something he had often advised in his professional life. He vowed to prioritize their relationship, no longer letting work or other distractions come between them.
\\
\\
As the afternoon sun cast long shadows across the room, Lina leaned against Gabriel, their hearts beating in unison. They knew it wouldn't be easy, but with honesty, understanding, and a commitment to rebuilding their connection, they were ready to face whatever challenges lay ahead. Together, they were determined to turn their newfound priorities into a stronger, more resilient love.
\\
\\
Gabriel sat by the window of his small apartment, the city lights flickering outside, a stark contrast to the turmoil brewing within. The incident, a fire sparked by an overlooked candle, had brought him face to face with the reality of apartment living. While his friends and neighbors had rallied around him, providing emotional support and helping with the aftermath, his landlord saw the situation differently. He was convinced that Gabriel's negligence had caused the blaze, and there was talk of compensation for the damages.
\\
\\
Despite the pressure, Gabriel maintained his composure. He knew he had been careless, but he also believed in fairness. When the landlord finally agreed to meet, Gabriel prepared himself for a difficult negotiation. He wanted to rectify the situation without sacrificing his own stability. As the meeting began, Gabriel's heart raced, but he remained steadfast, his voice steady as he presented his case.
\\
\\
"I understand the concerns, and I take responsibility for my actions," Gabriel said, his gaze unwavering. "But I hope we can find a middle ground that acknowledges the unforeseen circumstances and the efforts I've made to correct the situation."
\\
\\
The landlord, initially rigid, seemed to soften at Gabriel's mature approach. They eventually reached an agreement that was fair to both parties. Gabriel would cover a portion of the damages through a payment plan, while the landlord would cover the rest as an acknowledgment of the unforeseen nature of the accident.
\\
\\
With the compensation settled, a weight lifted from Gabriel's shoulders. The support from his friends and the community, along with his own resilience, had not only helped him navigate the crisis but also deepened his relationships. Lina, a neighbor who had been a source of strength, now looked at him with a newfound respect. Their conversations, once surface-level, had taken on a more profound meaning, as they both shared their fears and insecurities.
\\
\\
Gabriel's friends, too, had shown their loyalty through the ordeal. They had stood by him, offering a listening ear and a helping hand when he needed it most. This experience had not only tested Gabriel's character but had also forged stronger bonds between them.
\\
\\
As the chapter closed, the apartment that once represented a milestone and independence now seemed like a symbol of growth and resilience. Gabriel, having faced adversity and come out the other side, felt a renewed sense of purpose. The urge to drown in self-blame had been replaced by a determination to learn from his mistakes and embrace the connections he had forged. It was a new beginning, one marked by understanding, forgiveness, and the promise of a brighter future.
\\
\\
Gabriel and Lina sat on the worn couch in their small apartment, their hands entwined as they gazed into each other's eyes. The room was dimly lit by a single lamp, casting warm shadows on the walls. They had been through so much together, and in that moment, they both knew their commitment to each other was stronger than ever.
\\
\\
"I want us to build a life based on love and understanding," Gabriel said softly, his voice filled with conviction. Lina smiled, her eyes shimmering with tears of joy. They had found solace in each other's embrace, and their deepening love was a beacon of hope in the midst of life's challenges.
\\
\\
Gabriel's next step was to give back, and he started a support group for others who were struggling with the same issues he had. In a small community center, he shared his experiences, the lessons he'd learned, and the importance of vulnerability. The group dynamic was a mix of pain, hope, and resilience, as strangers became friends, united by their shared struggles.
\\
\\
Each week, the group would gather, their stories weaving a tapestry of human experience. There was John, a recovering alcoholic, who found strength in the support of others. There was Maria, who grappled with the loss of her job and the fear of the unknown. Their friendships grew, providing a sense of belonging and purpose that was invaluable to all.
\\
\\
As Gabriel's involvement in the support group grew, so did his relationship with his family. They saw the changes in him, the newfound responsibility and growth. His parents, who had once been worried, now expressed pride in their son's transformation. Gabriel's sister, Sarah, reached out to him more often, their bond deepening as they talked about their shared experiences.
\\
\\
Gabriel's apartment, once a symbol of his struggles, now felt like a haven of progress. The walls that once echoed with anxiety now housed laughter and meaningful conversations. The fire safety issues he vowed to address were slowly being rectified, and the relationship with Lina, which had been through its own trials, was on the mend.
\\
\\
With each passing day, Gabriel felt a new beginning taking shape. The road ahead was still rocky, but he was no longer walking alone. He had Lina, his friends from the support group, and a family that was slowly healing. Embracing the challenges, he knew that together, they would forge a future filled with love, understanding, and growth.
\\
\\
Gabriel sat across from Lina in their cozy apartment, the afternoon sun casting warm shadows across the room. The tension that had been hanging in the air for days seemed to have reached its peak, and Gabriel knew it was time to confront any remaining issues head-on. He took a deep breath and spoke softly, "Lina, I want you to know that I've been thinking a lot about our conversations, and I understand now how my anxiety has affected us both."
\\
\\
Lina's trembling hands stilled as she looked into Gabriel's sincere eyes. She felt a wave of relief wash over her, grateful for his newfound maturity and willingness to address their concerns. "Gabriel, I'm glad you're seeing things clearly now. It's been hard being apart, and I've missed you so much," she admitted, her voice steadier than before.
\\
\\
Together, they decided to tackle a specific challenge that had been causing strain in their relationship – the distance between them due to their demanding jobs. They brainstormed ways to bridge the gap, taking turns to suggest ideas and compromises. Their teamwork was evident, and they both felt a renewed sense of unity as they worked through the issue.
\\
\\
Word had spread throughout their community about the challenges Gabriel and Lina were facing, and their friends and neighbors had rallied around them. They organized a small gathering, inviting the couple to share their experiences and struggles. The outpouring of love and support was overwhelming, and Gabriel and Lina felt a deep sense of belonging in the embrace of their community.
\\
\\
As the evening drew to a close, Gabriel and Lina left the gathering, hand in hand. The air was crisp, but their hearts were warm with the knowledge that they had not only overcome personal obstacles but had also strengthened their bond as a team. They knew that there would be more challenges to face, but with honesty, open communication, and the support of their friends, they were ready to face whatever life had in store for them.
\\
\\
The sun shone brightly on the small community park, casting a warm glow over the gathering of smiling faces. A celebration of unity was underway, organized to honor the individuals who had brought hope and healing to their neighborhood. At the heart of it all was Gabriel's support group, a place where people like him found solace and strength in their shared struggles.
\\
\\
Gabriel, standing alongside Lina, felt a rush of gratitude for the unexpected kindness that had enveloped his life. As the event progressed, he couldn't help but steal glances at Lina, whose eyes shimmered with the same joy that had ignited their renewed sense of unity. Her presence alone was a testament to their growth and the love they had built together.
\\
\\
In a touching moment, Gabriel's landlord, Mr. Thompson, approached him with a heartfelt gesture. He presented Gabriel with a small potted plant, symbolizing the growth and nurturing of their once-strained relationship. Gabriel was taken aback by the unexpected show of appreciation, his heart swelling with gratitude for the understanding and support he had received from those around him.
\\
\\
As the day wore on, Gabriel and Lina found themselves alone on a bench, their hands entwined. They shared a quiet moment, their gazes locked, understanding each other without words. The challenges they had faced together had forged a bond that was unbreakable, and in that instant, they knew they were ready to face whatever lay ahead.
\\
\\
With the park's sounds fading into the background, Gabriel whispered to Lina about their future, their plans to take the next steps together. They talked about the possibility of moving to a bigger place, somewhere they could build a life that was both comfortable and filled with love. Lina's eyes sparkled with anticipation, her tears of joy now replaced with a determined glint.
\\
\\
As the sun began to set, casting a golden hue over the park, Gabriel and Lina stood up, hand in hand. United and resolute, they faced the horizon, ready to embrace the challenges and the opportunities that life had in store. Their hearts filled with gratitude, they knew that together, they could conquer anything.
\\
\\
\bottomrule
\caption{The example of generated story.}
\label{tab: story example1}
\end{longtable}
\end{small}